\documentclass[letterpaper]{article}
\usepackage{aaai2026}  % DO NOT CHANGE THIS
\usepackage{times}  % DO NOT CHANGE THIS
\usepackage{helvet}  % DO NOT CHANGE THIS
\usepackage{courier}  % DO NOT CHANGE THIS
\usepackage[hyphens]{url}  % DO NOT CHANGE THIS
\usepackage{graphicx} % DO NOT CHANGE THIS
\usepackage{natbib}  % DO NOT CHANGE THIS AND DO NOT ADD ANY OPTIONS TO IT
\usepackage{caption} % DO NOT CHANGE THIS AND DO NOT ADD ANY OPTIONS TO IT
\usepackage{amsmath}
\usepackage{amsfonts}
\usepackage{booktabs}
\usepackage{nicefrac}
\usepackage{microtype}
\usepackage{subcaption}
\usepackage{algorithm}
\usepackage{algorithmic}

\title{Forecasting N-Body Dynamics: A Comparative Study of Neural Ordinary Differential Equations and Universal Differential Equations}
\author{
    Suriya R S, Prathamesh Dinesh Joshi, Rajat Dandekar, Raj Dandekar, Sreedath Panat
}
\affiliations{
    Vizuara AI Labs
}

\begin{document}

\maketitle

\begin{abstract}
The n-body problem, fundamental to astrophysics, simulates the motion of n bodies acting under the effect of their own mutual gravitational interactions. Traditional machine learning models that are used for predicting and forecasting trajectories are often data-intensive "black box" models, which ignore the physical laws, thereby lacking interpretability. Whereas Scientific Machine Learning ( Scientific ML ) directly embeds the known physical laws into the machine learning framework. Through robust modelling in the Julia programming language, our method uses the Scientific ML frameworks: Neural ordinary differential equations (NODEs) and Universal differential equations (UDEs) to predict and forecast the system's dynamics. In addition, an essential component of our analysis involves determining the "forecasting breakdown point", which is the smallest possible amount of training data our models need to predict future, unseen data accurately. We employ synthetically created noisy data to simulate real-world observational limitations. Our findings indicate that the UDE model is much more data efficient, needing only 20\% of data for a correct forecast, whereas the Neural ODE requires 90\%. 
\end{abstract}

\section{Introduction}
The classical n-body problem in astrophysics seeks to predict the motion of a system of celestial objects that interact gravitationally with each other. Although an analytical closed-form solution exists for a system of two objects, no such solution has been discovered for three or more objects. As a result, historically, numerical integration methods such as the Runge-Kutta method or leap-frog schemes have been used to simulate solutions. However, traditional solvers operate under the assumption that the physical model of an n-body system is perfectly known and complete. Therefore, this assumption limits our ability to apply it to a realistic scenario where the system might be subject to unmodeled physics.

To address these challenges, Scientific Machine Learning ( Scientific ML ) has emerged as a powerful paradigm where we shift our objective from just simulating a known physical model to discovering or correcting the governing equations directly from observational data. Scientific ML combines the expressive power of neural networks with the interpretability of differential equations. This approach has been successfully implemented in various scientific disciplines like fluid mechanics, circuit modelling, optics, gene expression, quantum circuits, and epidemiology \cite{baker2019workshop, dandekar2020machine, dandekar2020safe, abhijit2022new, ji2022autonomous, bills2020universal, lai2021structural, nieves2024uncertainty, wang2023hybridizing, ramadhan2024data, rackauckascapturing, sharma2023reviewA, sharma2023reviewB, aboelyazeed2023differentiable}.

Primarily, the progress in Scientific ML is driven by the following two frameworks: \textbf{Neural Ordinary Differential Equations} (NeuralODEs) \cite{chen2018neural, dupont2019augmented, massaroli2020dissecting, yan2019robustness}, which learns the entire system dynamics through Neural Networks from data, and \textbf{Universal Differential Equations} (UDEs)\cite{rackauckas2020universal, bolibar2023universal, teshima2020universal, bournez2020universal}, which blends in the known physical laws with neural networks to learn only the unknown/unmodelled dynamics from data. While these frameworks are being used in astrophysics \cite{gupta2022galaxy,branca2023neural,origer2024closing}, a thorough comparative analysis of their effectiveness in solving problems is yet to be determined. In this study, we try to understand the effectiveness and limitations of these two Scientific ML frameworks.

Specifically, we aim to address the following questions in the context of the n-body problem:
\begin{enumerate}
    \item Can the UDE framework be used to learn and recover the pairwise gravitational interaction term by replacing it with a neural network?
    \item How does the predictive accuracy of NeuralODEs compare to that of UDEs when modelling the trajectories?
    \item Can both NeuralODEs and UDEs be used to forecast the system’s trajectories in the long term?
    \item Do UDEs, incorporating known physics, offer superior performance in forecasting over the purely data-driven NeuralODEs?
\end{enumerate}

We perform rigorous comparative analysis using advanced Scientific ML libraries to answer these questions. Our work provides critical insights into the effectiveness and limitations of these frameworks. Furthermore, we analyze the forecasting breakdown point as a metric to quantify the time horizon beyond which long-term predictions diverge from actual dynamics.

\section{Methodology}

The foundation of our study uses the classical Newtonian formulation of the gravitational N-body problem. For a system of n-bodies, the state evolution of the system is described by their positions $r_i(t) \in \mathbb{R}^3$ and velocities $v_i(t) \in \mathbb{R}^3$ of each body $i$. The state evolution form a system of ordinary differential equations ( ODEs ) given by, 

\begin{equation}
    \frac{dr_i}{dt} = v_i
    \label{eq:position_ode}
\end{equation}

\begin{equation}
    \frac{dv_i}{dt} = G \sum_{j \neq i} m_j \frac{r_j - r_i}{\|r_j - r_i\|^3}
    \label{eq:velocity_ode}
\end{equation}

Where $G$ is the gravitational constant and $m_j$ is the mass of body j. This system of equations is used for generating ground truth data and as the structural prior for the UDE model. 

\paragraph{Dataset generation}
A stable 3-body system was simulated with Runge–Kutta by Tsitouras' numerical integrator in the Julia Programming language for dataset generation. The simulation was run for 7 seconds, the domain $t \in [0, 10]$ was discretised into 70 equally spaced time points. The position and velocity at each of these time instants were saved. We create three distinct training datasets based on the noise level added. The first is the noise-free dataset that we get from the numerical simulation. Second is a moderate noise dataset with a Gaussian noise with a standard deviation of 7\% of the data range. Third, it is a high-noise dataset, where Gaussian noise with a standard deviation of 35\% of the data range is added. The noise is primarily added to simulate real-world synthetic data.

We use 100\% of the dataset for the prediction task for training. For forecasting, the models were first trained on an initial portion of the dataset, with the remaining portion used for evaluation. More specifically, we used training-forecasting splits of 90\%-10\%, 80\%-20\%, 40\%-60\%, and 20\%-80\% of the time points.

\subsection{Neural Ordinary Differential Equation (NODE)}
In this approach, we define the dynamics of the system's hidden state vector $ h(t) $ with the help of an ordinary differential equation where the function describing the change is a neural network $f$ parameterised by $\theta$. Here, we perform backpropagation through the neural network augmented ODE. In doing so, we find the optimal values of the neural network parameters. 

\begin{equation}
    \frac{dh}{dt} = f(h(t), t, \theta)
\end{equation}

In our application to the n-body problem, we consider the state vector $h(t) \in \mathbb{R}^{3n}$ which is just the concatenation of the position vectors and the velocity vectors of all the n-bodies $h(t) = [r_1, v_1, r_2, v_2, \dots, r_n, v_n]$. Here, the Neural ODE framework replaces all the dynamics (Equations~\eqref{eq:position_ode} and \eqref{eq:velocity_ode}) of the n-body system with a neural network. By training on the generated data, the network could learn the complex gravitational dynamics of the entire system without prior knowledge of the physics. The model's hyperparameters were selected after searching a grid of possible values specific to the training data used.

\subsection{Universal Differential Equations (UDEs)}

In contrast to NeuralODEs, UDEs offer a hybrid approach where only the specific, unknown, or incomplete terms of an ODE/PDE system are learned by replacing them with a neural network. This allows for correcting existing physical models and the data-driven discovery of new physical principles. 

In the context of the n-body problem, UDE is formed by retaining the knowledge that the total acceleration of the body is the sum of pairwise interactions with all other bodies. In addition, we replace the interaction term itself with a neural network. This allows the neural network to discover the underlying gravitational interaction between the objects from the data. The UDE is therefore defined as 

\begin{equation}
    \frac{dr_i}{dt} = v_i
\end{equation}
\begin{equation}
    \frac{dv_i}{dt} = \sum_{j=1, j \neq i}^{n} \mathrm{NN}(r_i, r_j, m_i, m_j, \theta)
\end{equation}

Where $\theta$ denotes the parameters of the Neural Network (NN), the input to the network is the states (positions $r_i$, $r_j$, and masses $m_i$, $m_j$) of the two interacting bodies $i$  and $j$. In this configuration, we assume the kinematic relationship and summation structure are already known physical laws. The neural network is learning the core interaction between any two bodies. Just as with the Neural ODE, the hyperparameters for the UDE model had been selected from exhaustive searches that corresponded to the dataset used in training.

\section{Results}
We have considered a total of 5 cases with different dataset percentages, evaluating each of them under the following three noise levels: no noise, moderate noise(7\% standard deviation), and high noise (35\% standard deviation). The main paper presents the results from training the NeuralODE and UDE models on the complete dataset and on the data subsets of 80\% and 20\%. Here, we also include the analysis of the forecasting breakdown point. Results for the remaining cases, which utilized 90\% (Case 2) and 40\% (Case 4) of the data for training, are located in Appendix A.

\subsection{Case 1: Training on complete dataset}

\begin{figure}[t!]
    \centering
    \begin{subfigure}[b]{0.32\columnwidth}
        \includegraphics[width=\linewidth]{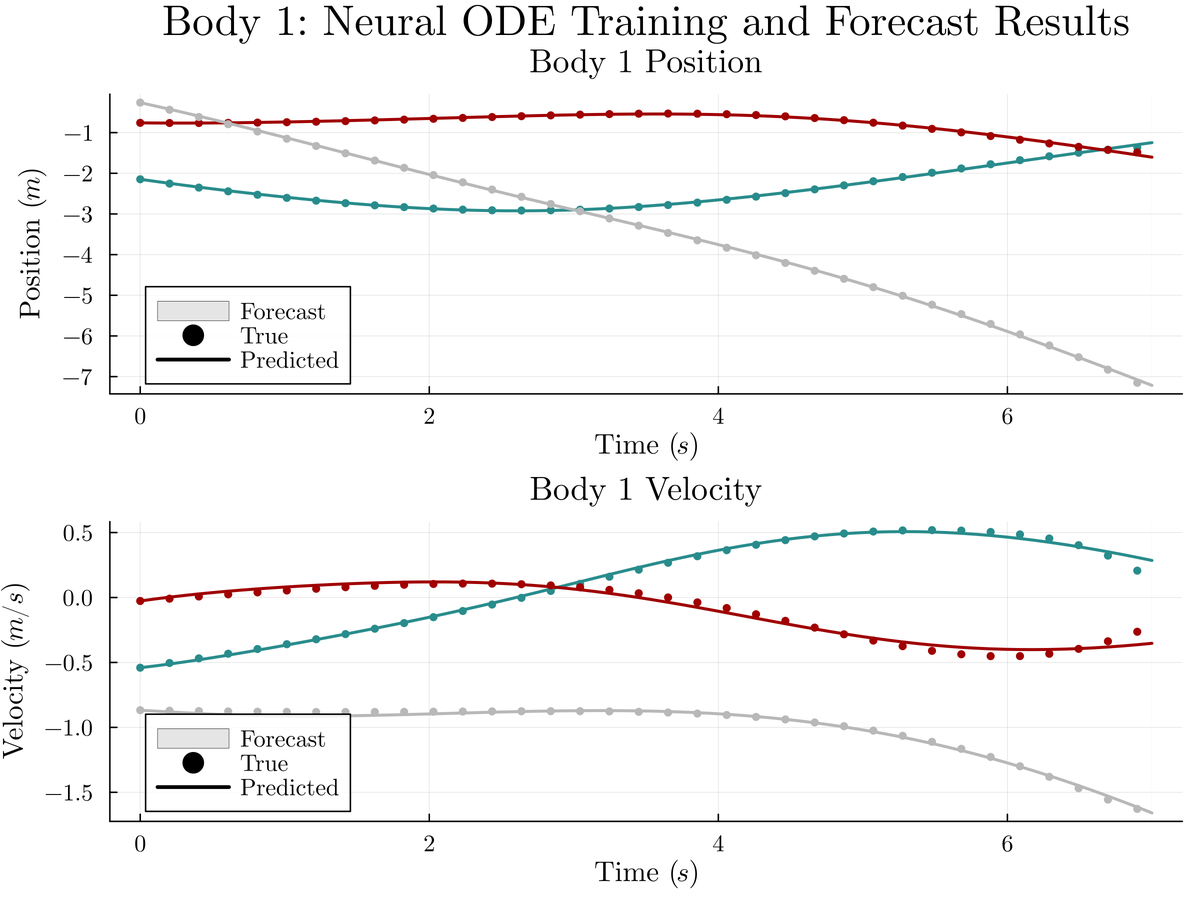}
        \caption{No noise}
    \end{subfigure}
    \hfill
    \begin{subfigure}[b]{0.32\columnwidth}
        \includegraphics[width=\linewidth]{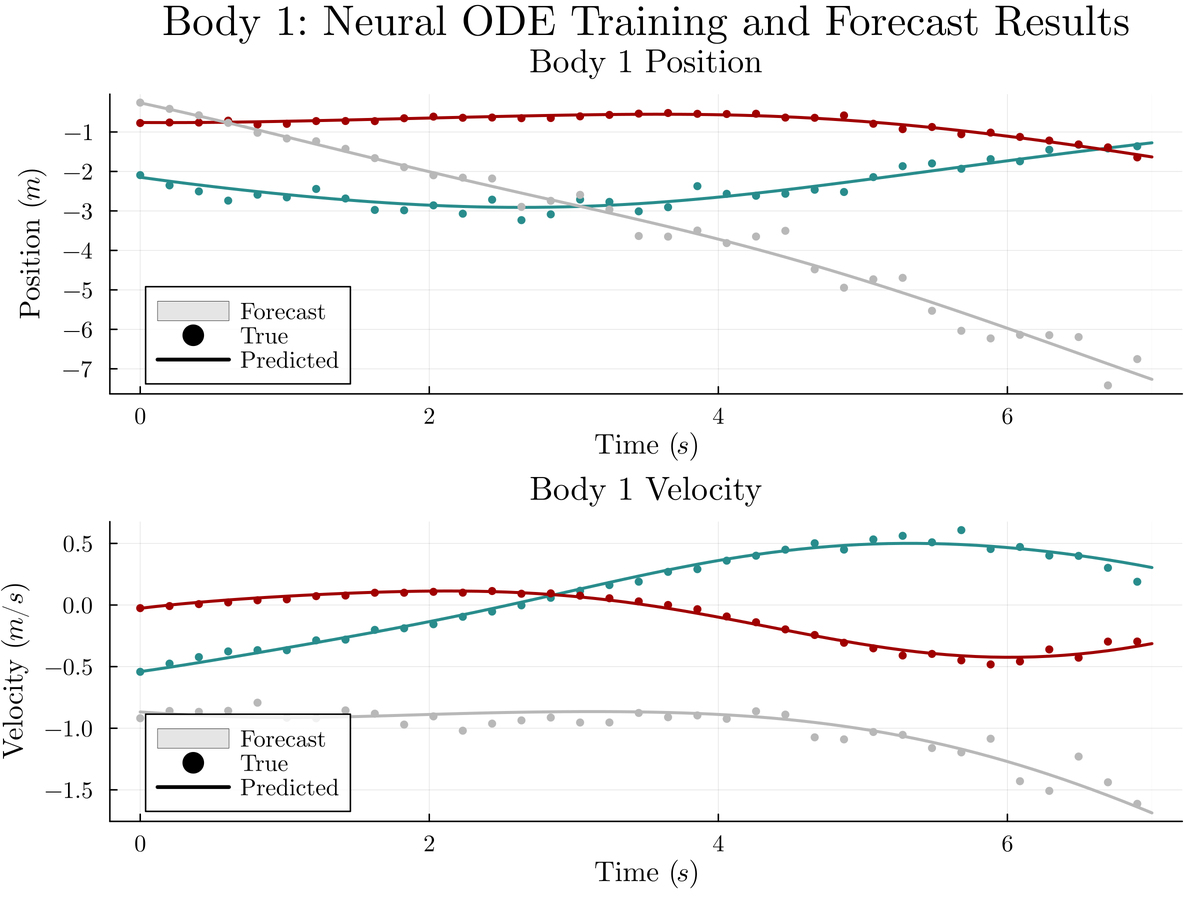}
        \caption{Moderate noise}
    \end{subfigure}
    \hfill
    \begin{subfigure}[b]{0.32\columnwidth}
        \includegraphics[width=\linewidth]{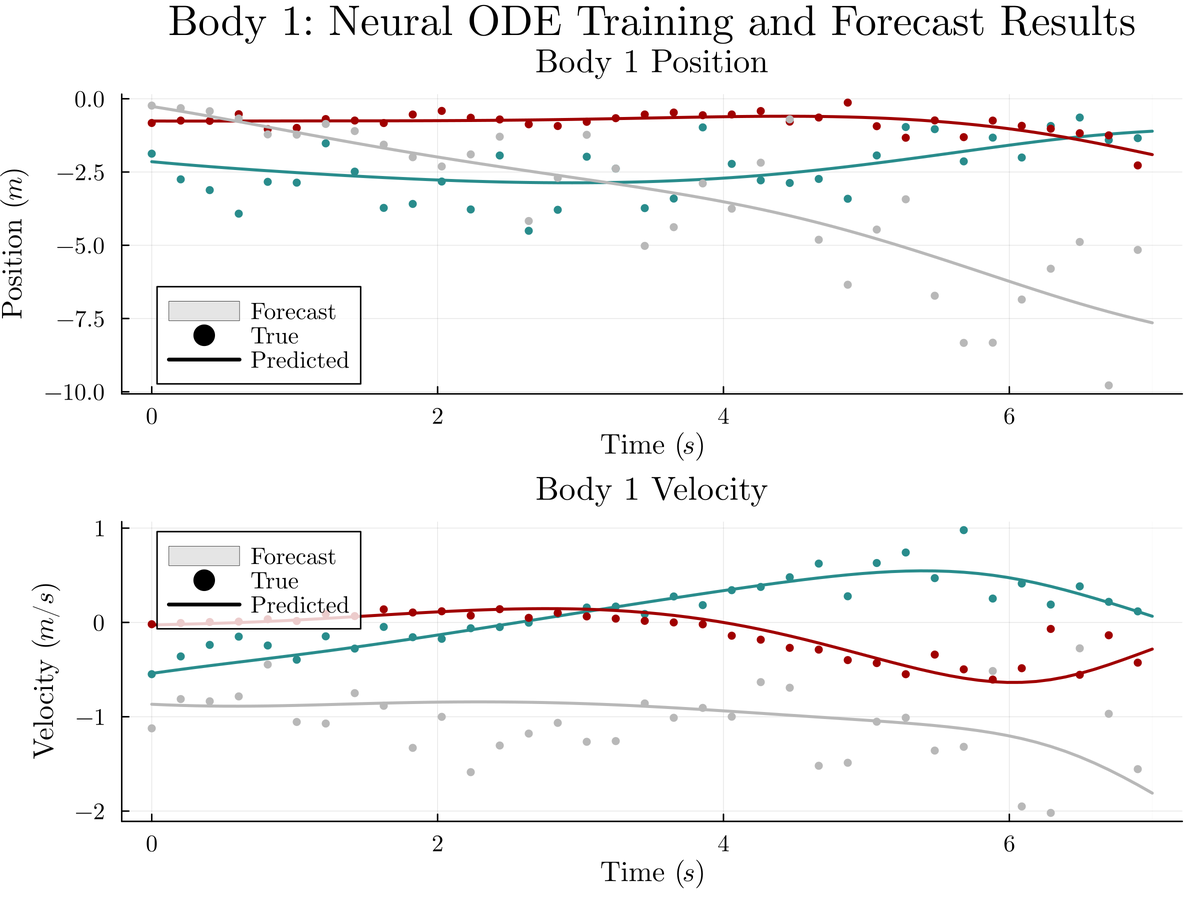}
        \caption{High noise}
    \end{subfigure}
    \caption{Neural ODE results for Case 1 (100\% training) across different noise levels for body 1.}
    \label{fig:case1_node}
\end{figure}

From Figure \ref{fig:case1_node}, it is evident that the Neural ODE effectively learns the n-body dynamics throughout the entire time span and across different noise conditions. When trained on noise-free data, its forecasts for the position and velocity of each body closely align with the actual trajectories. As noise is introduced, the model persists in generating smooth and physically plausible trajectories, successfully filtering out a majority of the random fluctuations inherent in the training data. Notably, this level of performance is maintained even as noise intensifies, with the model reliably producing accurate and stable trajectories. These findings underscore the model's considerable robustness to noise, ensuring long-term positional accuracy even in scenarios with significant corruption.

\begin{figure}[t!]
    \centering
    \begin{subfigure}[b]{0.32\columnwidth}
        \includegraphics[width=\linewidth]{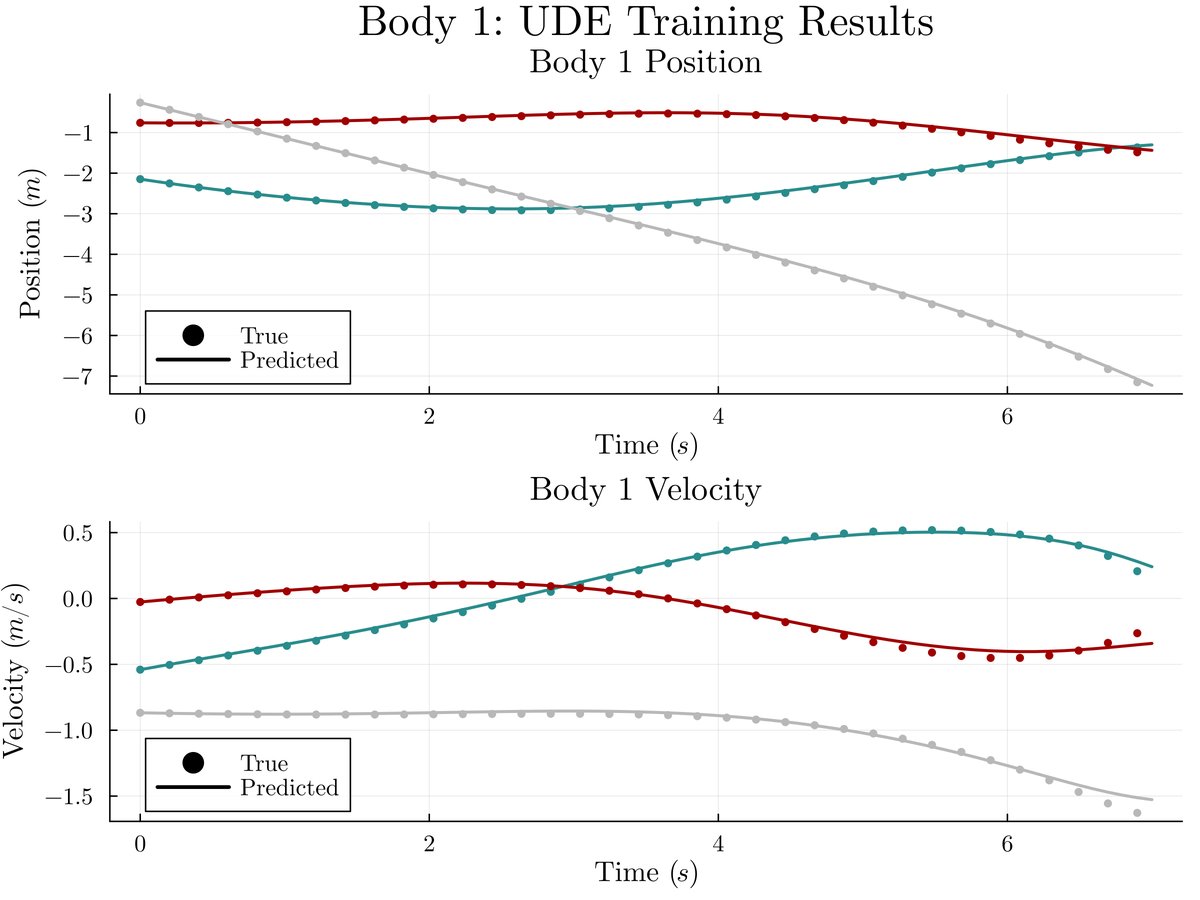}
        \caption{No noise}
    \end{subfigure}
    \hfill
    \begin{subfigure}[b]{0.32\columnwidth}
        \includegraphics[width=\linewidth]{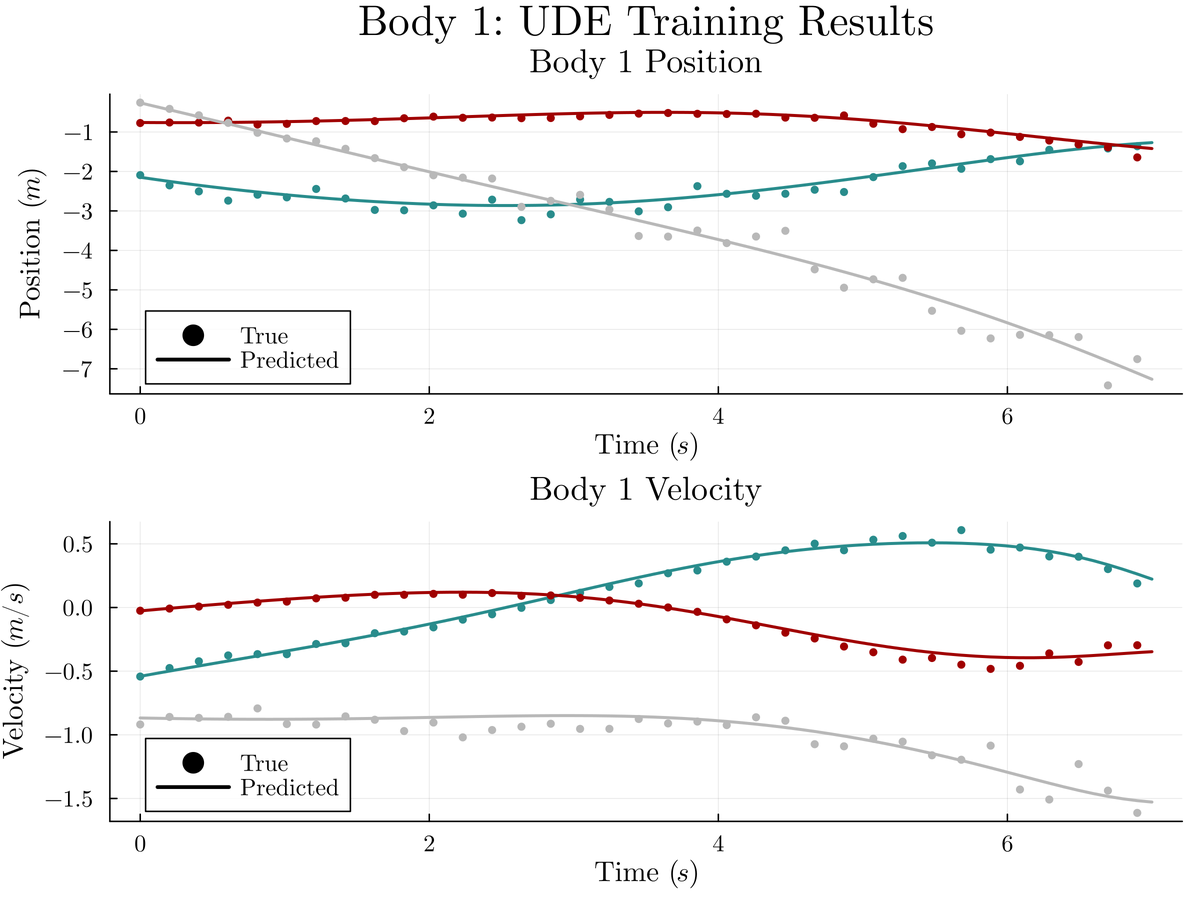}
        \caption{Moderate noise}
    \end{subfigure}
    \hfill
    \begin{subfigure}[b]{0.32\columnwidth}
        \includegraphics[width=\linewidth]{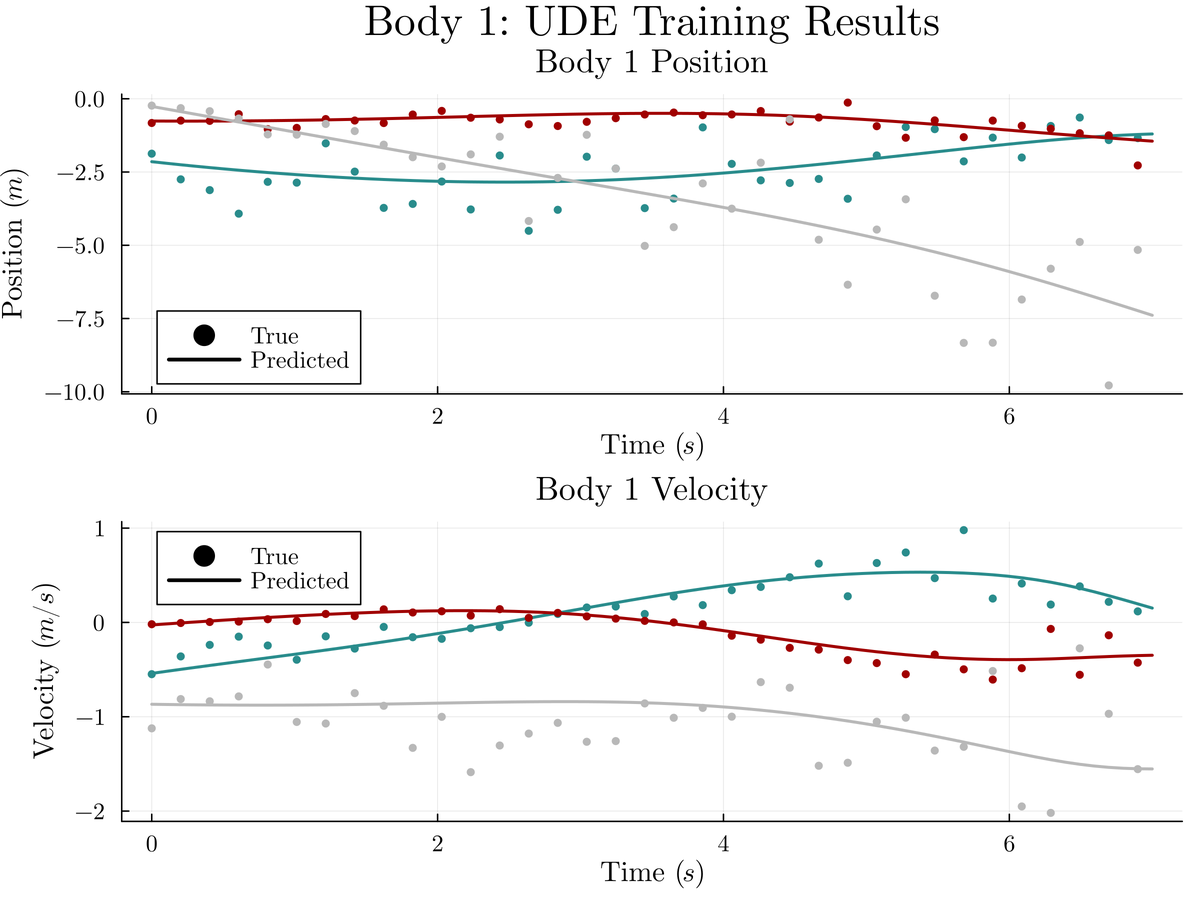}
        \caption{High noise}
    \end{subfigure}
    \caption{UDE results for Case 1 (100\% training) across different noise levels for body 1.}
    \label{fig:case1_ude}
\end{figure}

From Figure \ref{fig:case1_ude}, the UDE, which is trained on the entire dataset, performs remarkably well in predicting the 3-body trajectories in all the noise levels. When the dataset is noise-free, throughout the duration of the simulation, the model’s trajectory perfectly aligns with the ground truth data. Even under the moderate noise, the UDE produces clean and accurate predictions following the underlying dynamics. Even though under high noise, the data points have become significantly scattered, the model’s predictions remain smooth and physically realizable. While minor deviations from the true path may appear, the overall shape and evolution of the trajectory are preserved. These results demonstrate that when given access to the full dataset, the UDE is highly effective at learning the correct system dynamics and robust to substantial noise.

\subsection{Case 3: Training on 80\% of the dataset and forecasting}
\begin{figure}[t!]
    \centering
    \begin{subfigure}[b]{0.32\columnwidth}
        \includegraphics[width=\linewidth]{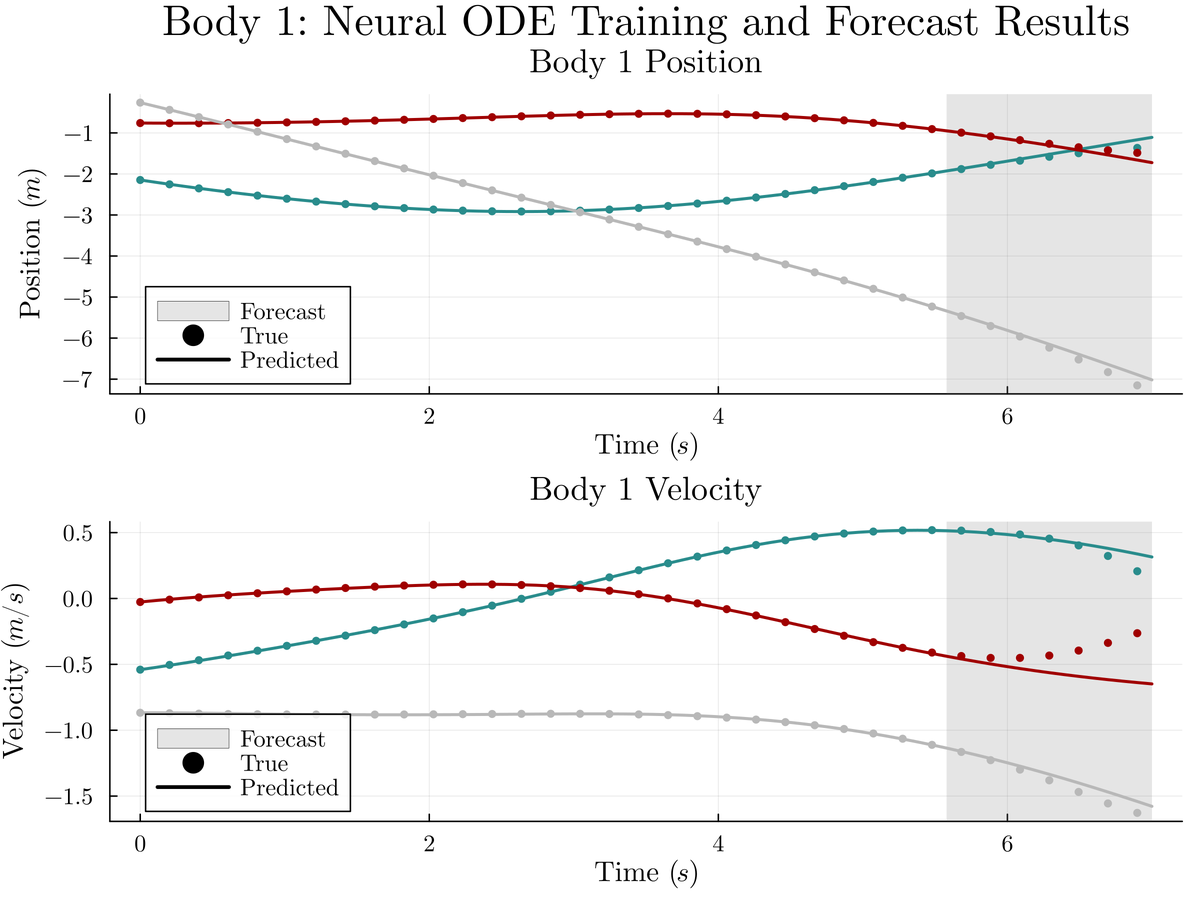}
        \caption{No noise}
    \end{subfigure}
    \hfill
    \begin{subfigure}[b]{0.32\columnwidth}
        \includegraphics[width=\linewidth]{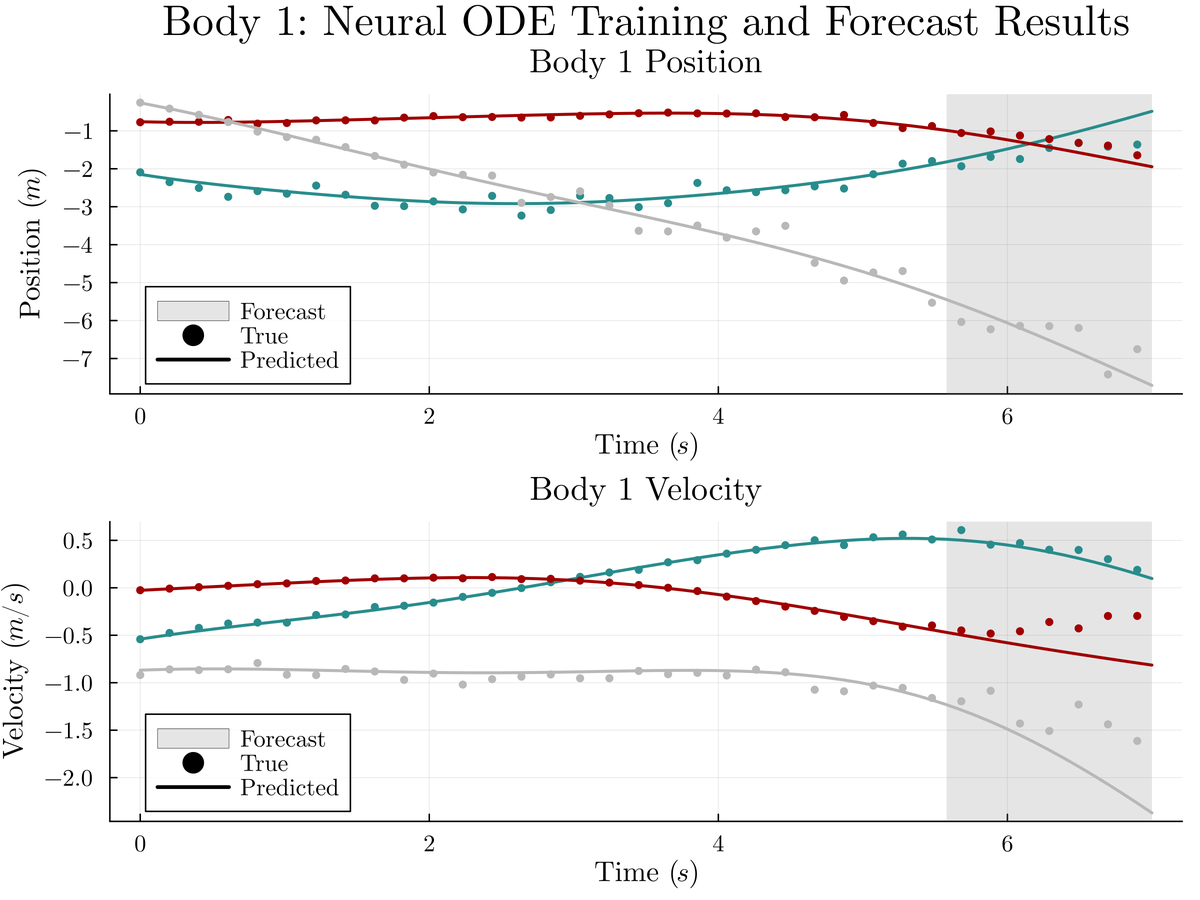}
        \caption{Moderate noise}
    \end{subfigure}
    \hfill
    \begin{subfigure}[b]{0.32\columnwidth}
        \includegraphics[width=\linewidth]{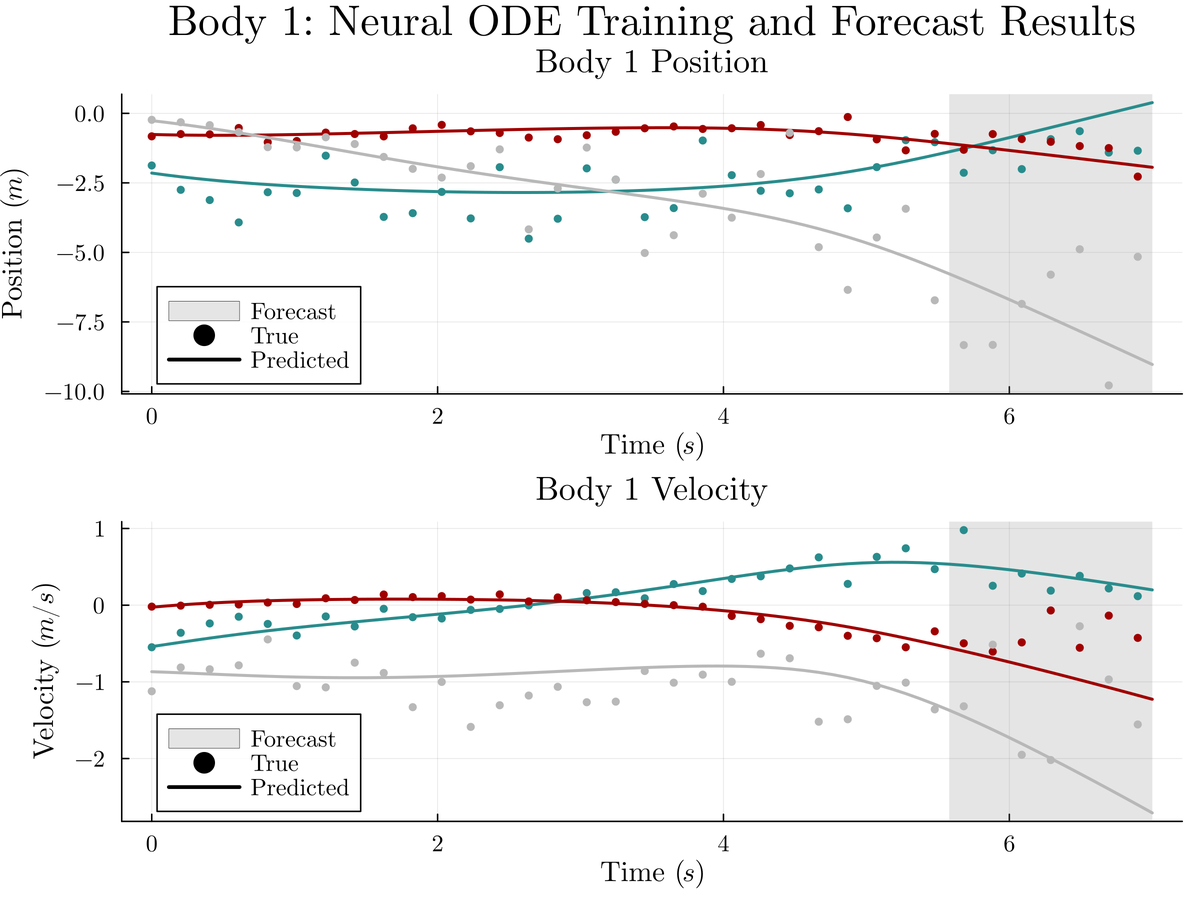}
        \caption{High noise}
    \end{subfigure}
    \caption{Neural ODE results for Case 3 (80\% training) across different noise levels for body 1.}
    \label{fig:case3_node}
\end{figure}

As depicted in Figure \ref{fig:case3_node}, wherein the Neural ODE is trained on 80\% of the temporal domain and subsequently predicts the remaining 20\%, a notable disparity in forecasting accuracy is observed between position and velocity. In the absence of noise, the predicted position trajectories align closely with the ground truth within both training and forecasting domains; conversely, the velocity predictions exhibit early signs of divergence. Under moderate noise conditions, the model continues to produce smooth and physically plausible position trajectories that adhere to the central trends of the noisy data; however, the accuracy of the velocity forecast deteriorates significantly, revealing considerable error. In scenarios with high noise levels, the training data becomes considerably more dispersed, and while the position outputs from the Neural ODE maintain a degree of smoothness, they exhibit a discernible decline in accuracy within the forecasting domain, with the velocity forecast completely failing. Ultimately, these findings underscore that the model's ability to generate reliable forecasts is compromised when it is trained using only 80\% of the available data. 

\begin{figure}[t!]
    \centering
    \begin{subfigure}[b]{0.32\columnwidth}
        \includegraphics[width=\linewidth]{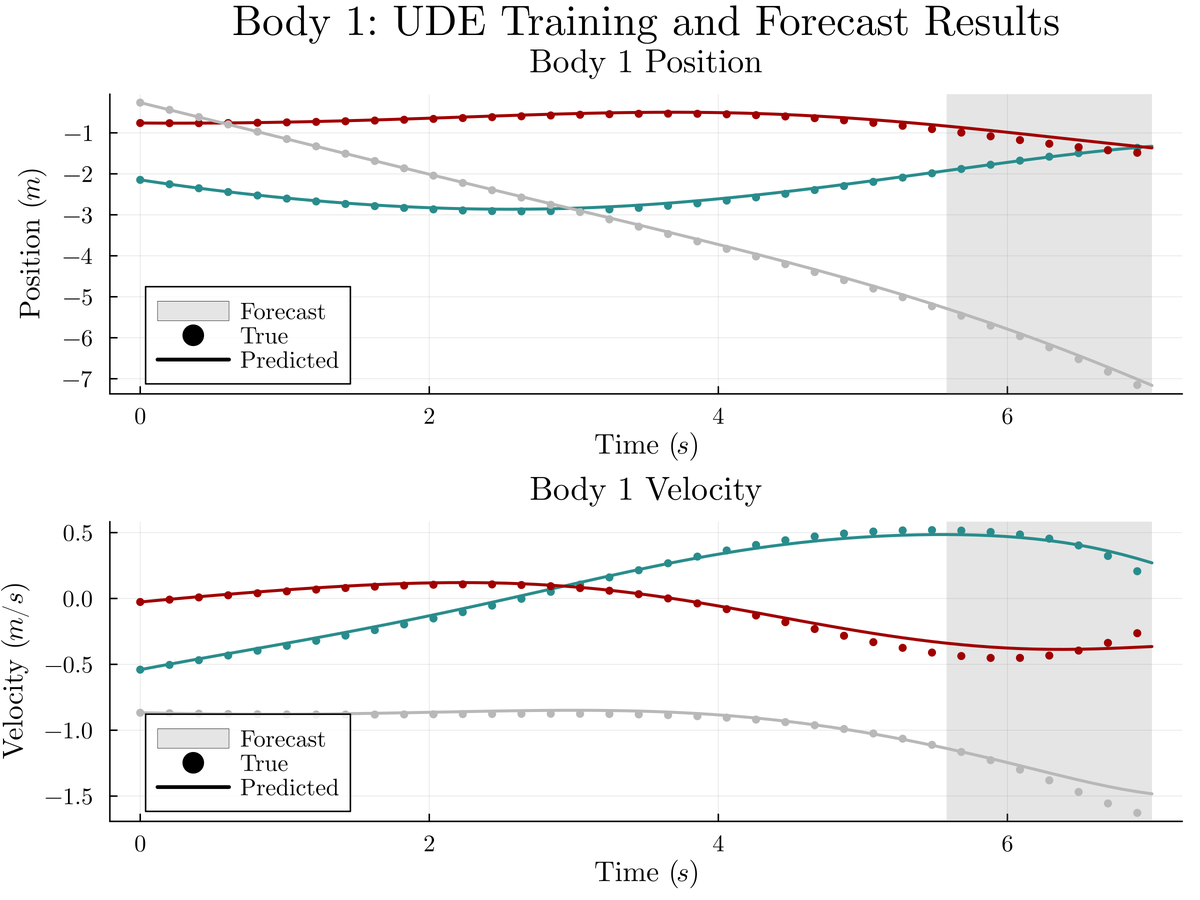}
        \caption{No noise}
    \end{subfigure}
    \hfill
    \begin{subfigure}[b]{0.32\columnwidth}
        \includegraphics[width=\linewidth]{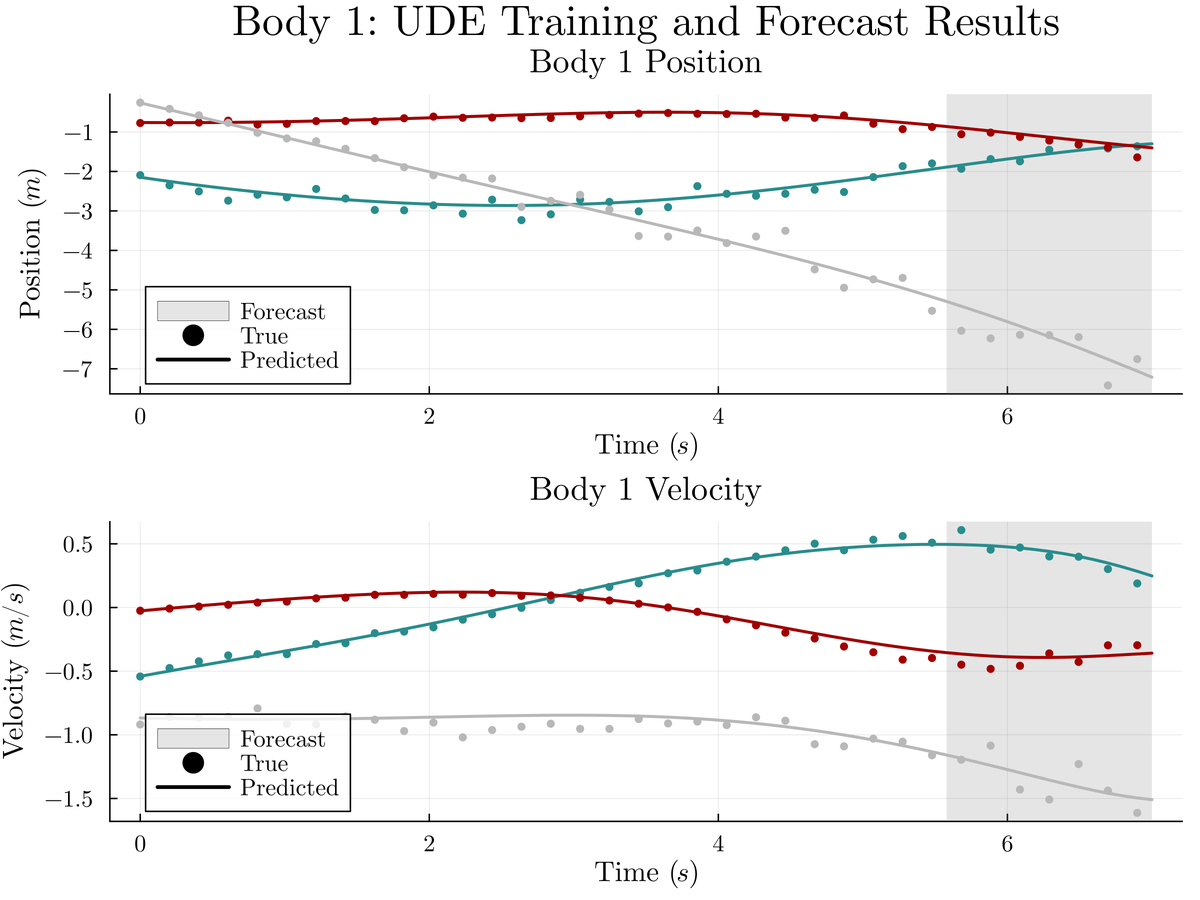}
        \caption{Moderate noise}
    \end{subfigure}
    \hfill
    \begin{subfigure}[b]{0.32\columnwidth}
        \includegraphics[width=\linewidth]{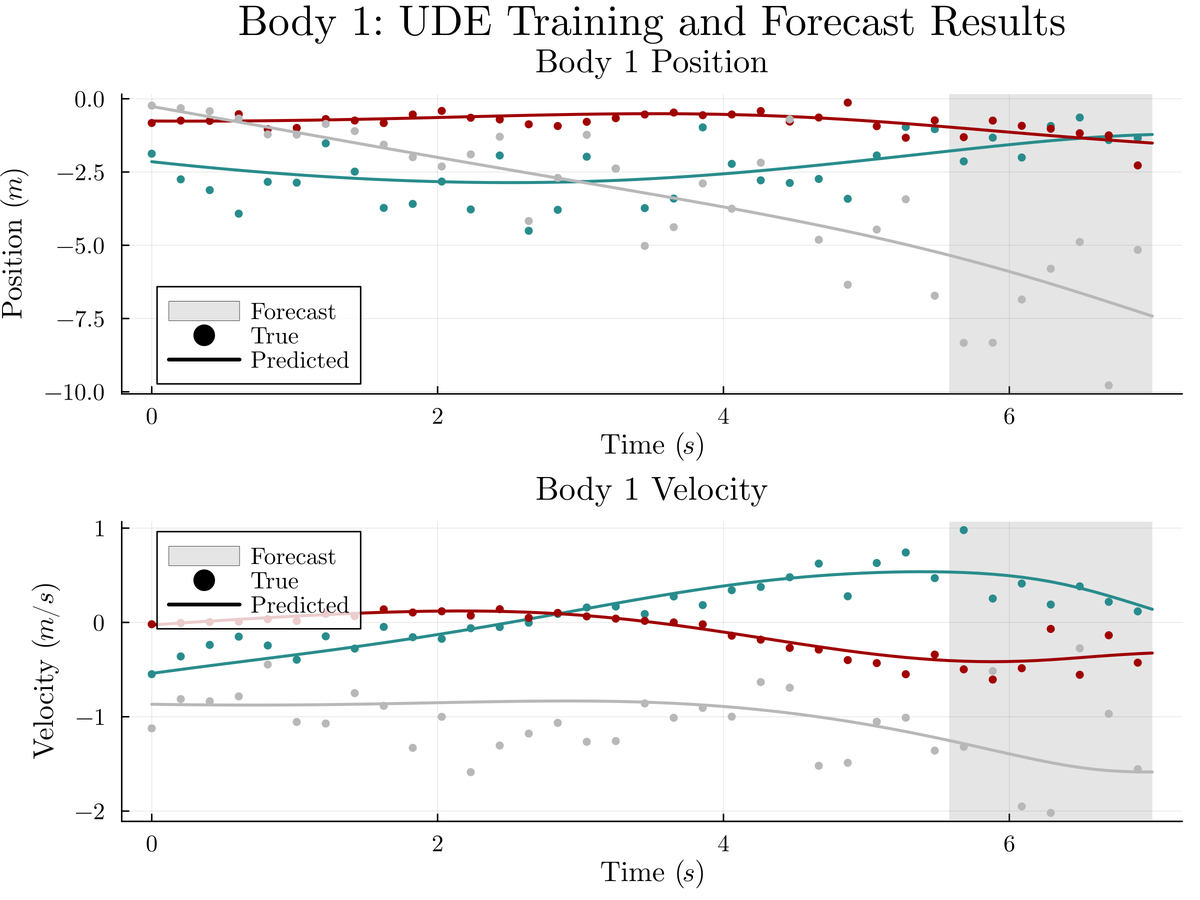}
        \caption{High noise}
    \end{subfigure}
    \caption{UDE results for Case 3 (80\% training) across different noise levels for body 1.}
    \label{fig:case3_ude}
\end{figure}

In Figure \ref{fig:case3_ude}, we have UDE trained on 80\% of the dataset and forecast on the rest. We can see it performs strongly across all noise levels. When the data is noise-free, its predictions for each body’s trajectory stay very close to the actual path, both during training and the short forecast interval, indicating excellent generalization. Under moderate noise, where the training points are a bit dispersed, the UDEs prediction remains smooth and follows the underlying dynamics very well. Under high noise, the UDEs still produce a clean forecast that stays close to the true trajectory.

\subsection{Case 5: Training on 20\% of the dataset and forecasting}
\begin{figure}[t!]
    \centering
    \begin{subfigure}[b]{0.32\columnwidth}
        \includegraphics[width=\linewidth]{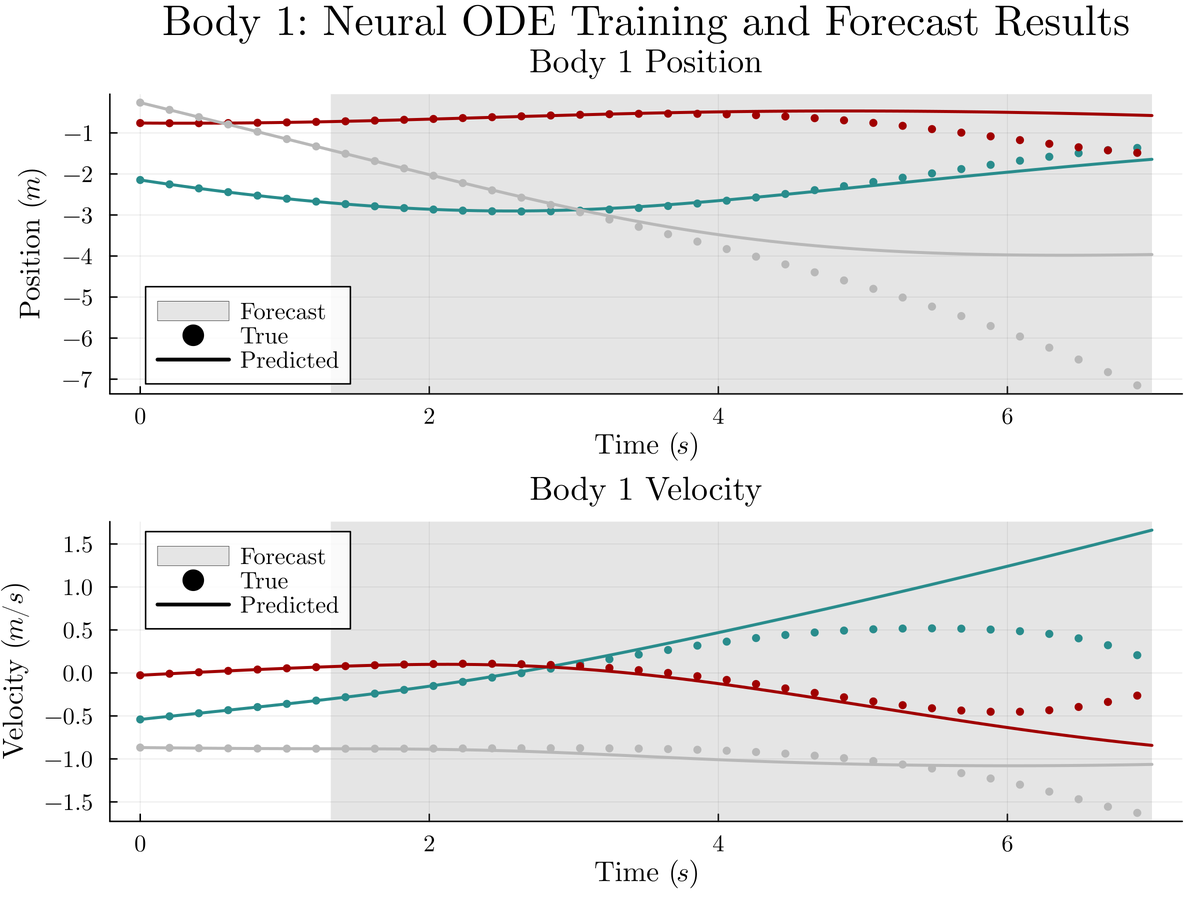}
        \caption{No noise}
    \end{subfigure}
    \hfill
    \begin{subfigure}[b]{0.32\columnwidth}
        \includegraphics[width=\linewidth]{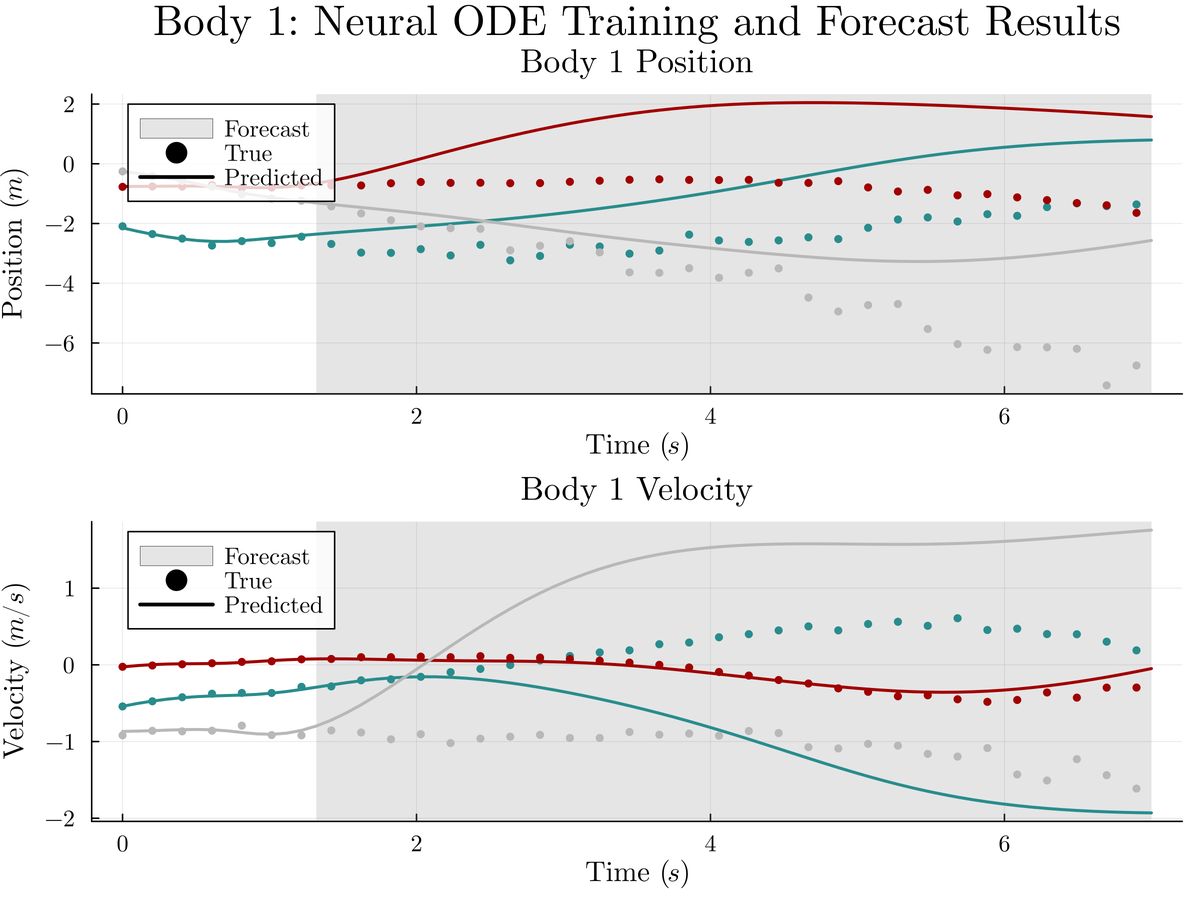}
        \caption{Moderate noise}
    \end{subfigure}
    \hfill
    \begin{subfigure}[b]{0.32\columnwidth}
        \includegraphics[width=\linewidth]{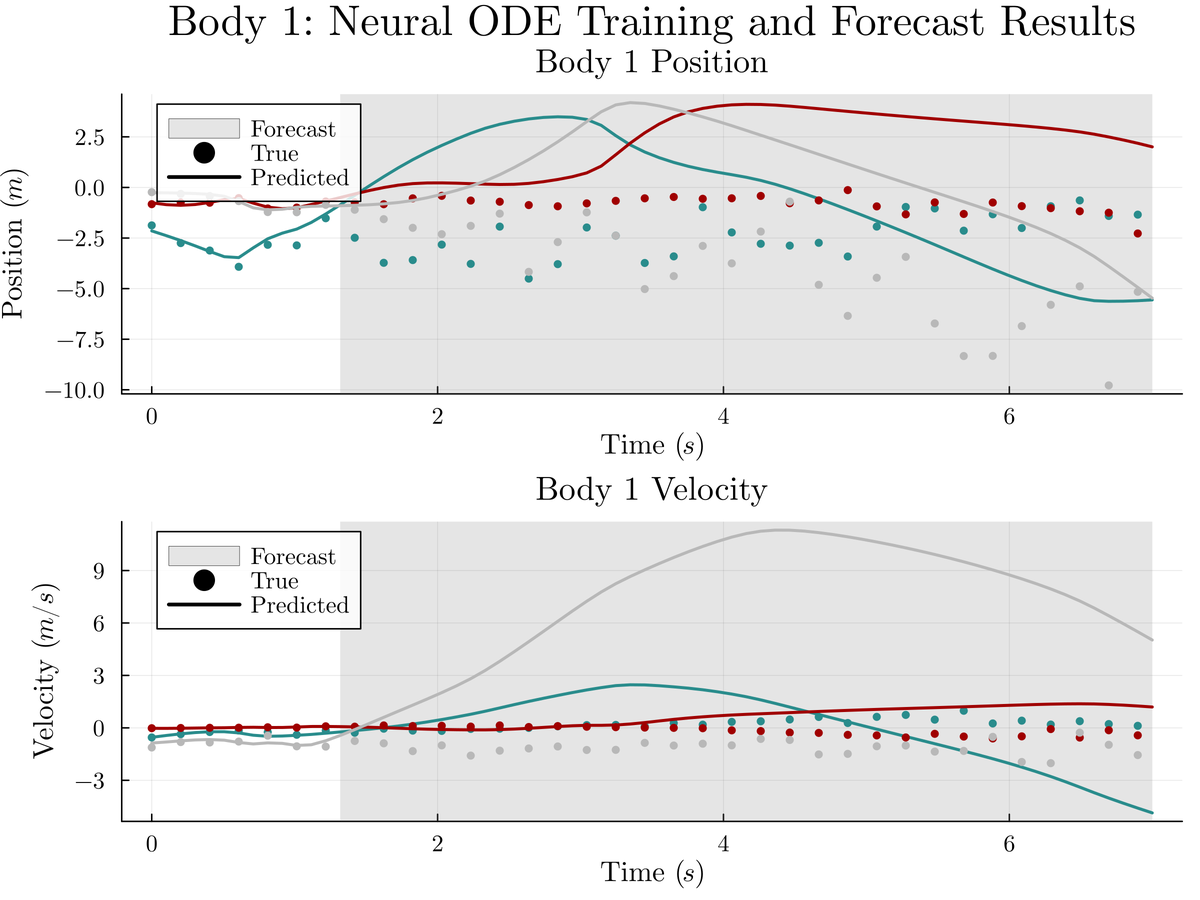}
        \caption{High noise}
    \end{subfigure}
    \caption{Neural ODE results for Case 5 (20\% training) across different noise levels for body 1.}
    \label{fig:case5_node}
\end{figure}

From Figure \ref{fig:case5_node}, when the Neural ODE is trained on only 20\% of the domain, its ability to forecast the remaining 80\% is significantly challenged. In the no-noise case, the predictions match the true paths well within the small training area, but the forecasted trajectories show clear and growing deviations over time. With moderate noise, the model captures the general trends within the limited training region; however, forecasting errors increase substantially, leading to predicted paths that diverge significantly from the ground truth. Under high noise, the model's forecasted trajectories lose coherence and show poor long-term predictive accuracy, failing to generalize from the sparse, noisy data.

\begin{figure}[h!]
    \centering
    \begin{subfigure}[b]{0.32\columnwidth}
        \includegraphics[width=\linewidth]{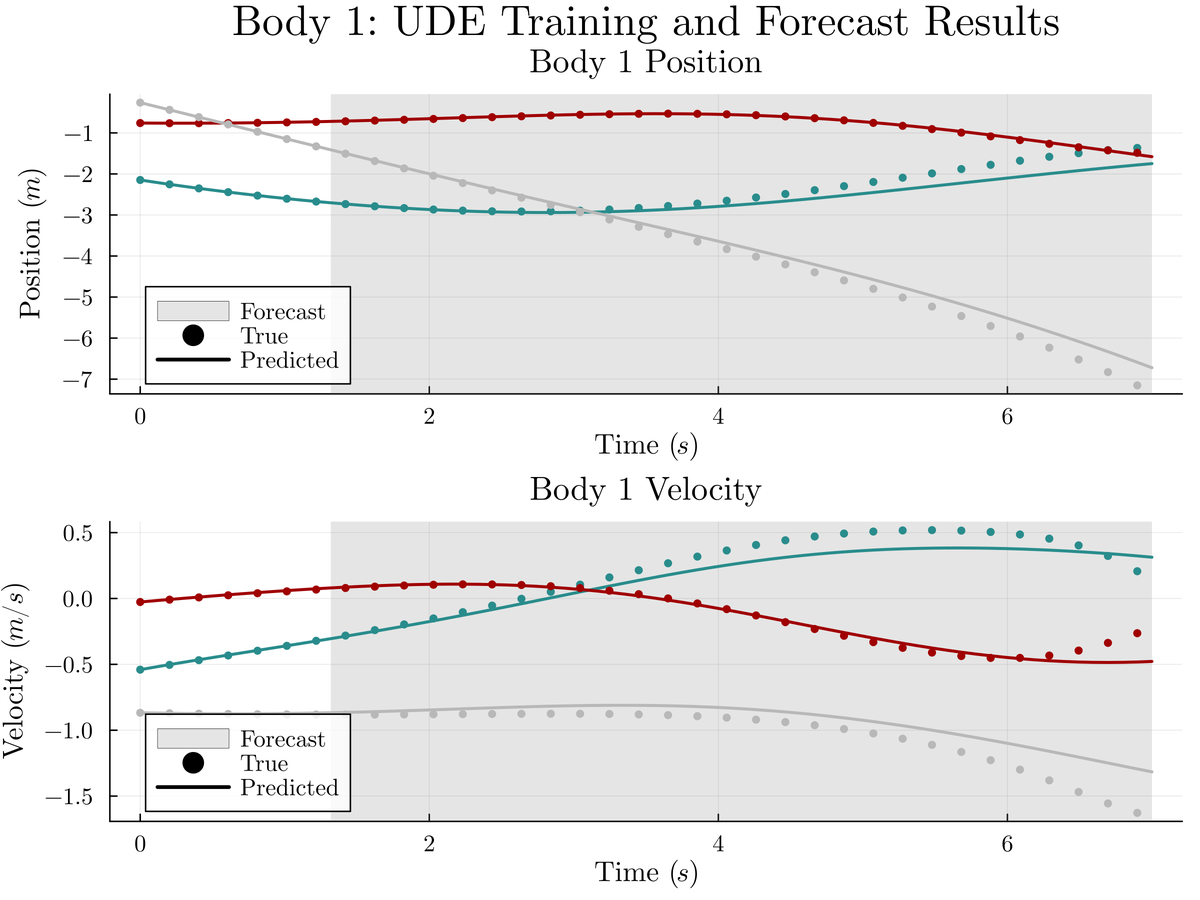}
        \caption{No noise}
    \end{subfigure}
    \hfill
    \begin{subfigure}[b]{0.32\columnwidth}
        \includegraphics[width=\linewidth]{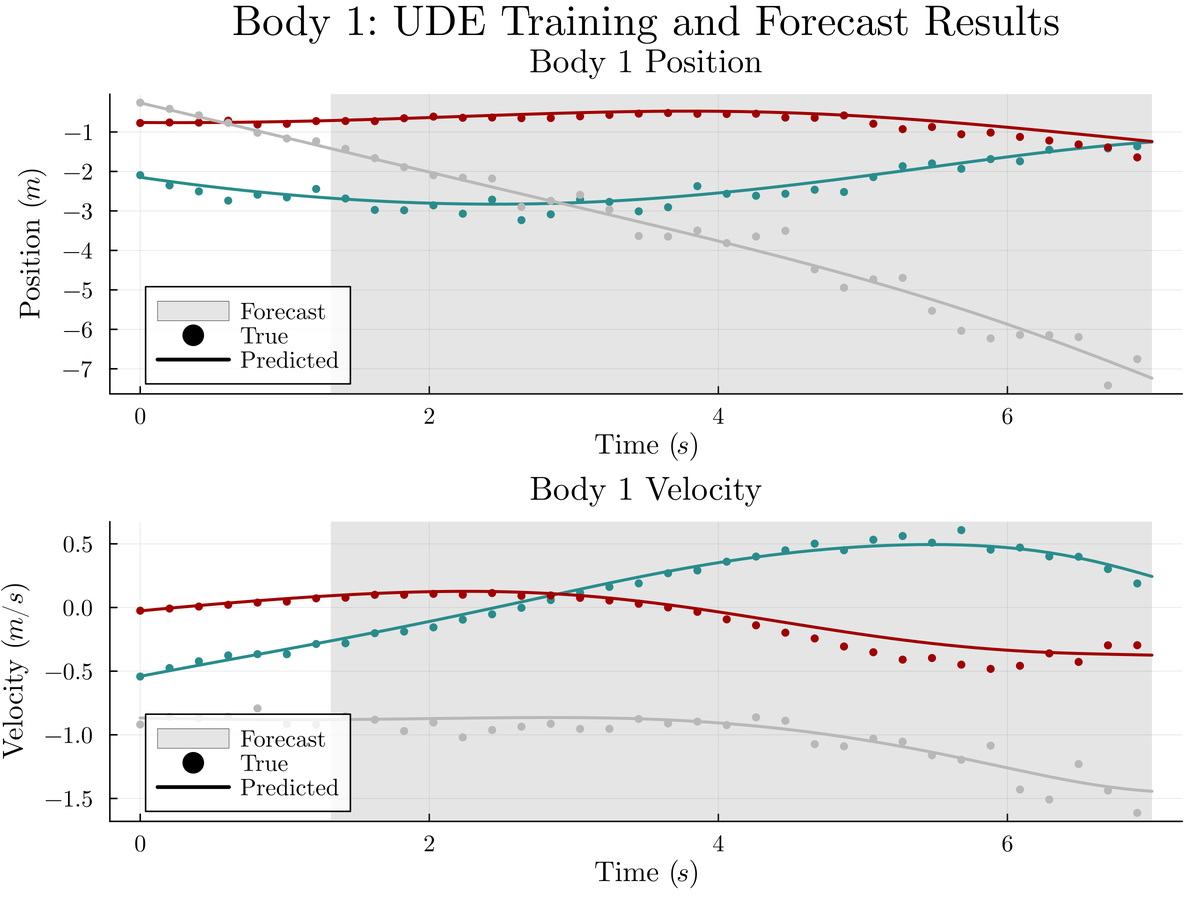}
        \caption{Moderate noise}
    \end{subfigure}
    \hfill
    \begin{subfigure}[b]{0.32\columnwidth}
        \includegraphics[width=\linewidth]{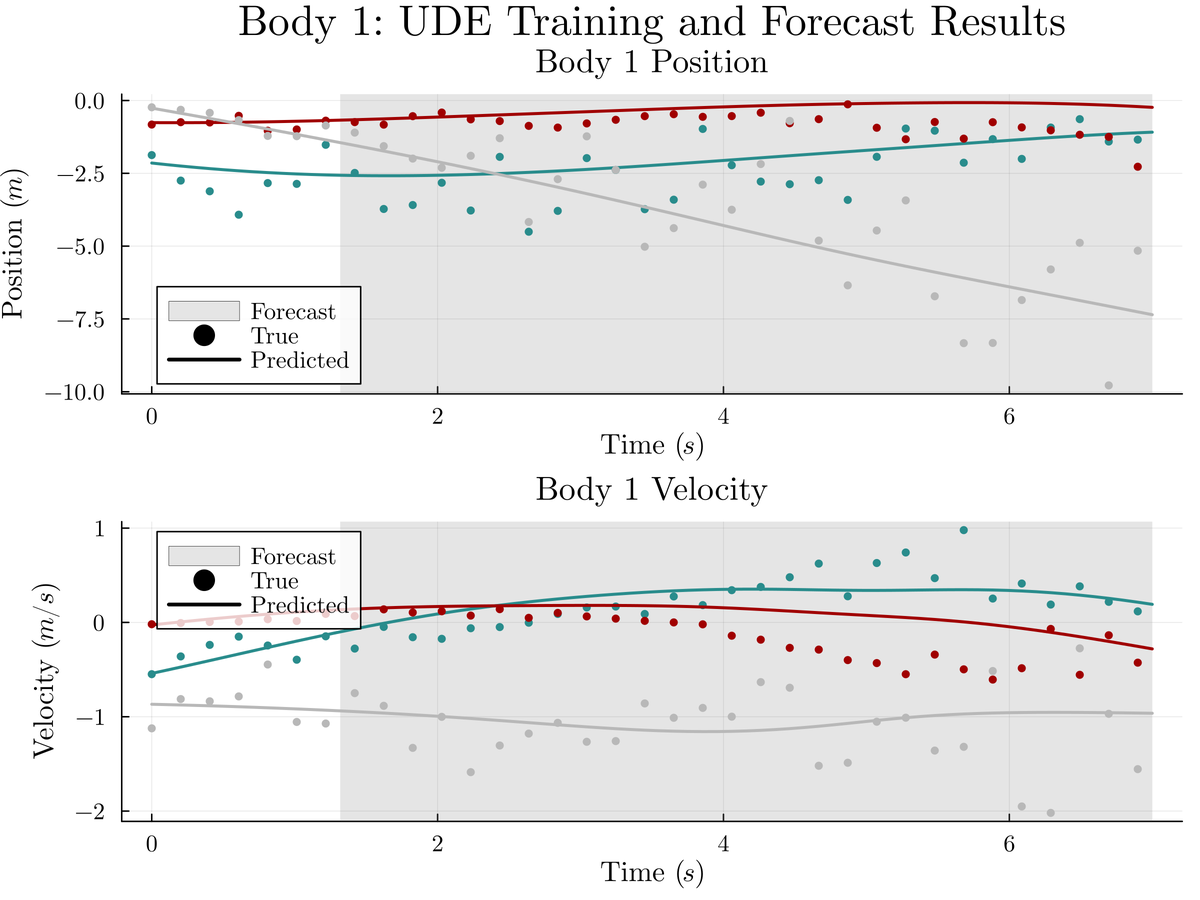}
        \caption{High noise}
    \end{subfigure}
    \caption{UDE results for Case 5 (20\% training) across different noise levels for body 1.}
    \label{fig:case5_ude}
\end{figure}

In Figure \ref{fig:case5_ude}, the UDE is trained on 20\% of the dataset and forecasted on the rest. Under a noise-free dataset, the UDE's prediction for each body's trajectory follows the true path almost exactly, not only within the limited training region but also far into the extended forecast area, demonstrating robust generalization from a small data subset. With moderate noise, the model still produces a reliable trajectory. Under high noise, although the graphs follow the trend, we can clearly see that the forecasting fails.

\subsection{Forecasting Breakdown Point Analysis}

To further explore the models' long-range forecasting capabilities, we progressively reduced the amount of training data to identify their forecasting breakdown points. This point is defined as the smallest percentage of training data below which the model fails to produce a physically plausible forecast of the unseen trajectory.

For the no-noise dataset, the Neural ODE required at least 90\% of the data, failing to forecast the future trajectory when trained on smaller subsets. In contrast, the UDE model demonstrated superior data efficiency, providing a reliable forecast even when trained with as little as 20\% of the available data. However, the UDE also failed when the training data was reduced to just 10\%. It is important to note that for the noisy datasets (moderate and high), both models required even larger data percentages to achieve stable forecasts.

\section{Discussion and Conclusion}
The paper here offers a comparative analysis between Neural Ordinary Differential Equations (Neural ODEs) and Universal Differential Equations (UDEs) for forecasting trajectories related to the gravity n-body problem under various data and noise setting. Neural ODEs had a strong ability to understand the system's dynamics when trained on the complete dataset. However, their forecasting success was highly dependent on data availability with more than 90\% dataset. Compared to this, UDEs were much more data-efficient, with low forecasting errors for models trained on as few as 20\% of the dataset. This data-robustness underscores the benefit of hardcoding known physical laws—the form of gravitational interactions in this example—into models. With only unknown or unmodeled factors to learn, the UDE formulation provides a more trustworthy route to generalization with few-data sets.

That being said, a few of its limitations hold for both models as well. For the 7-second simulation window, this work's concern lies with showing proof-of-concept recovery and short-term forecasting accuracy, not long-horizon stability. Numerous UDE and Neural ODE benchmarks employ shorter time intervals initially to guarantee local dynamics accuracy before moving to larger trajectories. Furthermore, this study was conducted using a single set of initial conditions, and the models' ability to learn the dynamics from different set of initial conditions haven't been explored. Finally, the analysis is confined to a 3-body system, leaving its scalability to systems with more bodies as an open question.  

These results are in agreement with the large-scale experience in the SciML community. For noisy, data-lean sitations, physics-informed models such as UDEs exhibit stronger generalizability compared to black-box models such as Neural ODEs. With the preserved structure of governing equations, UDEs exhibit increased interpretability, since learned neural components may be investigated for explaining discrepancies or unmodeled effects—an important benefit for scientific exploration. Future work will shift towards the long-term forecasting as well as generalizing the framework towards more complex gravity systems, for instance, with non-gravity forces or relativistic effects. The ultimate goal is to apply these Scientific ML models to real observational data, perhaps revealing new information about celestial mechanics and expanding beyond the reaches of our current physical models.

% --- IMPORTANT ---
% You must provide a .bib file for your references. 
% The name of the file should be 'references.bib' in this example.
% The aaai2026.sty file sets the bibliography style for you, so you don't need \bibliographystyle.
\bibliography{references}

%%%%%%%%%%%%%%%%%%%%%%%%%%%%%%%%%%%%%%%%%%%%%%%%%%%%%%%%%%%%
\appendix
\section{Appendix}

\subsection{Hyperparameter Details}

We determined the optimal hyperparameters for our models through a grid search. For the Neural ODE model, we employed a common two-stage optimization strategy: initial training with the Adam optimizer to quickly find a good region in the parameter space, followed by the BFGS optimizer for fine-tuning. The UDE model converged effectively using only the AdamW optimizer.

The tables below detail the search space for each key hyperparameter, with the final selected values shown in \textbf{bold}.

% --- TABLE FOR NEURAL ODE ---
\begin{table}[h!]
\centering
\caption{Neural ODE Hyperparameters.}
\label{tab:node_hyperparams}
\begin{tabular}{@{}ll@{}}
\toprule
\textbf{Hyperparameter} & \textbf{Value / Search Space} \\ \midrule
Activation Function   & ReLU, \textbf{tanh}, swish \\
Hidden Layers         & 2, \textbf{3}, 4 \\
Units per Layer       & 16, 32, \textbf{64}, 128 \\
Adam Learning Rate    & 1e-4, \textbf{1e-3}, 1e-2 \\
Adam Epochs           & 100, \textbf{200}, 500 \\
BFGS Epochs           & 100, \textbf{200}, 500 \\ \bottomrule
\end{tabular}
\end{table}

% --- TABLE FOR UDE ---
\begin{table}[h!]
\centering
\caption{UDE Hyperparameters.}
\label{tab:ude_hyperparams}
\begin{tabular}{@{}ll@{}}
\toprule
\textbf{Hyperparameter} & \textbf{Value / Search Space} \\ \midrule
Optimization Solver   & Adam, \textbf{AdamW} \\
Activation Function   & ReLU, tanh, \textbf{swish} \\
Hidden Layers         & \textbf{1}, 2, 3 \\
Units per Layer       & 16, \textbf{32}, 64 \\
Learning Rate         & 1e-4, \textbf{1e-3}, 1e-2 \\
Epochs                & 500, \textbf{700}, 1000 \\ \bottomrule
\end{tabular}
\end{table}

\subsection{Additional Forecasting Results}
This appendix presents the two additional cases (90\% and 40\% training coverage) omitted from the main text for brevity. Both follow the same experimental setup and evaluation protocol described in Section 2.

\subsubsection{Case 2: Training on 90\% of dataset and forecasting}
In this case, the models were trained on 90\% of the time domain and evaluated on the remaining 10\%. As this meets the minimum data requirement for the Neural ODE, its performance is stable, though still sensitive to noise.

\begin{figure}[h!]
    \centering
    \begin{subfigure}[b]{0.32\columnwidth}
        \includegraphics[width=\linewidth]{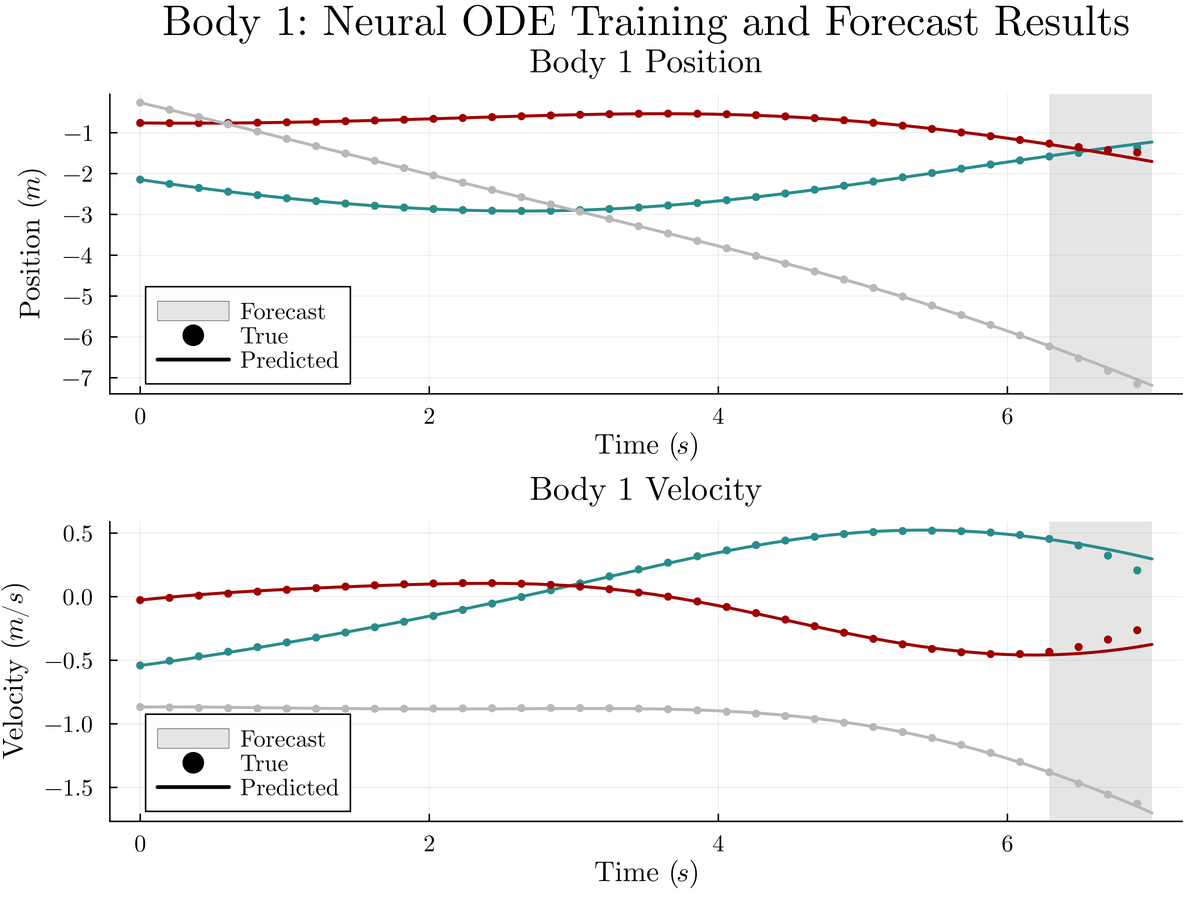}
        \caption{No noise}
    \end{subfigure}
    \hfill
    \begin{subfigure}[b]{0.32\columnwidth}
        \includegraphics[width=\linewidth]{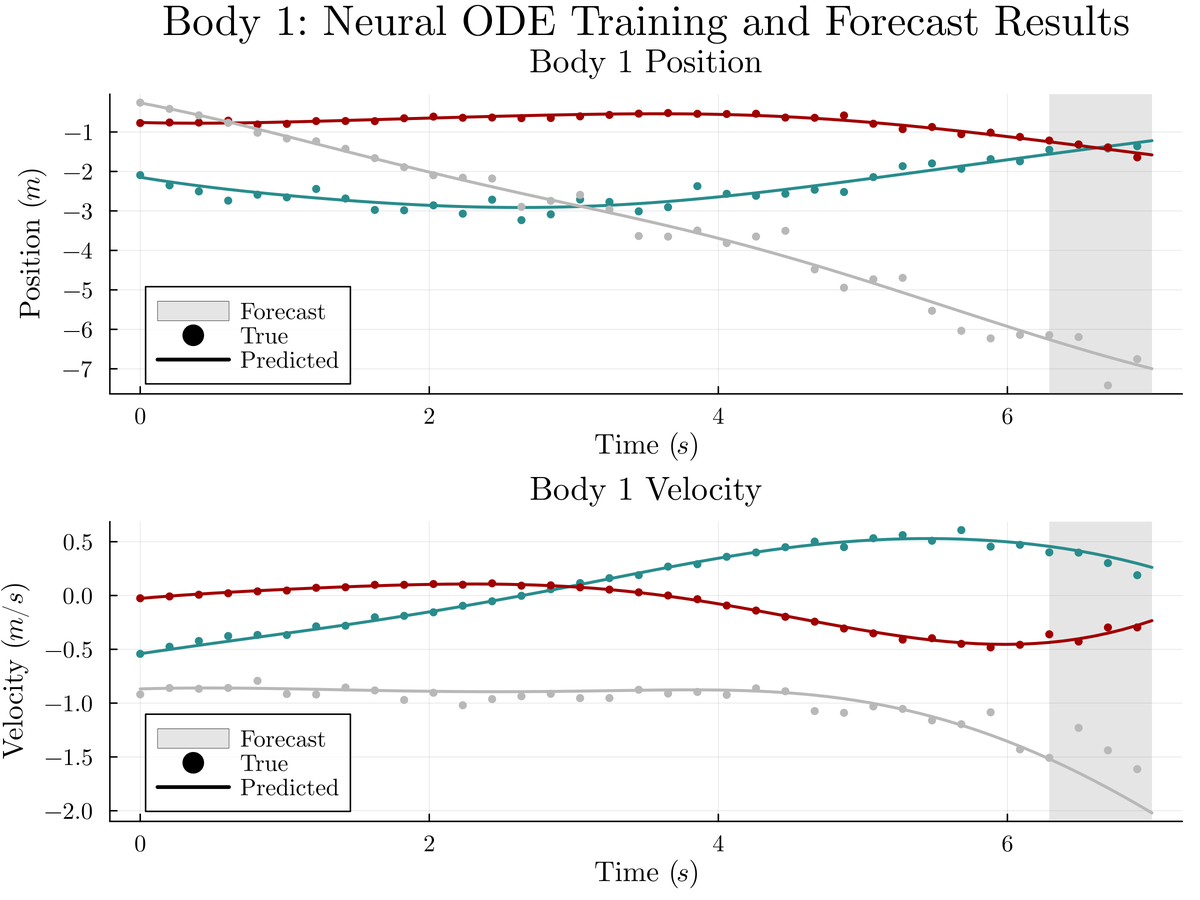}
        \caption{Moderate noise}
    \end{subfigure}
    \hfill
    \begin{subfigure}[b]{0.32\columnwidth}
        \includegraphics[width=\linewidth]{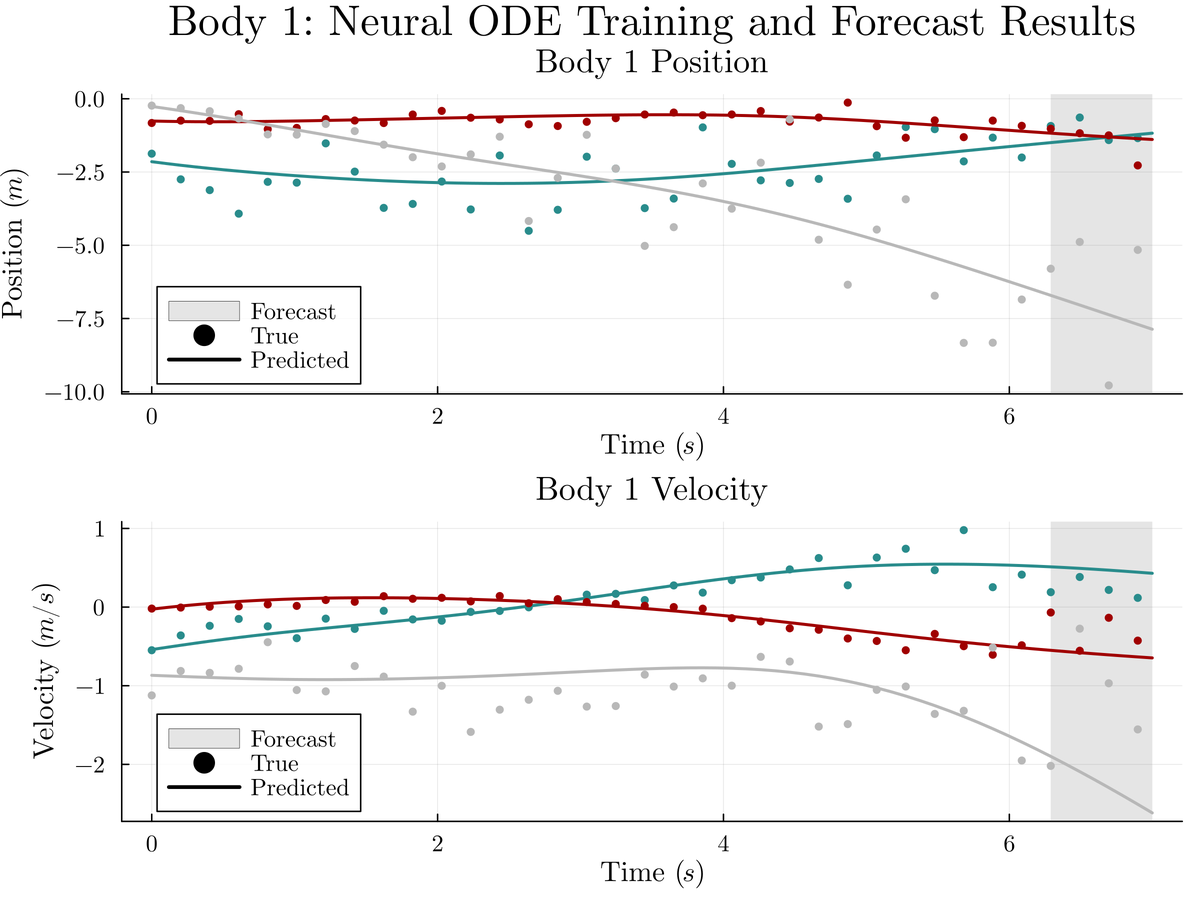}
        \caption{High noise}
    \end{subfigure}
    \caption{Neural ODE results for Case 2 (90\% training) across different noise levels for body 1.}
    \label{fig:case2_node_app}
\end{figure}

From Figure \ref{fig:case2_node_app}, the Neural ODE's predictions align well with the true trajectories in the no-noise scenario for both the training and the short 10\% forecast window. When noise is introduced, the model effectively filters it within the training region. However, minor deviations and phase shifts begin to appear in the forecast region, especially under high noise, indicating that even with substantial data, long-term stability is not guaranteed.

\begin{figure}[h!]
    \centering
    \begin{subfigure}[b]{0.32\columnwidth}
        \includegraphics[width=\linewidth]{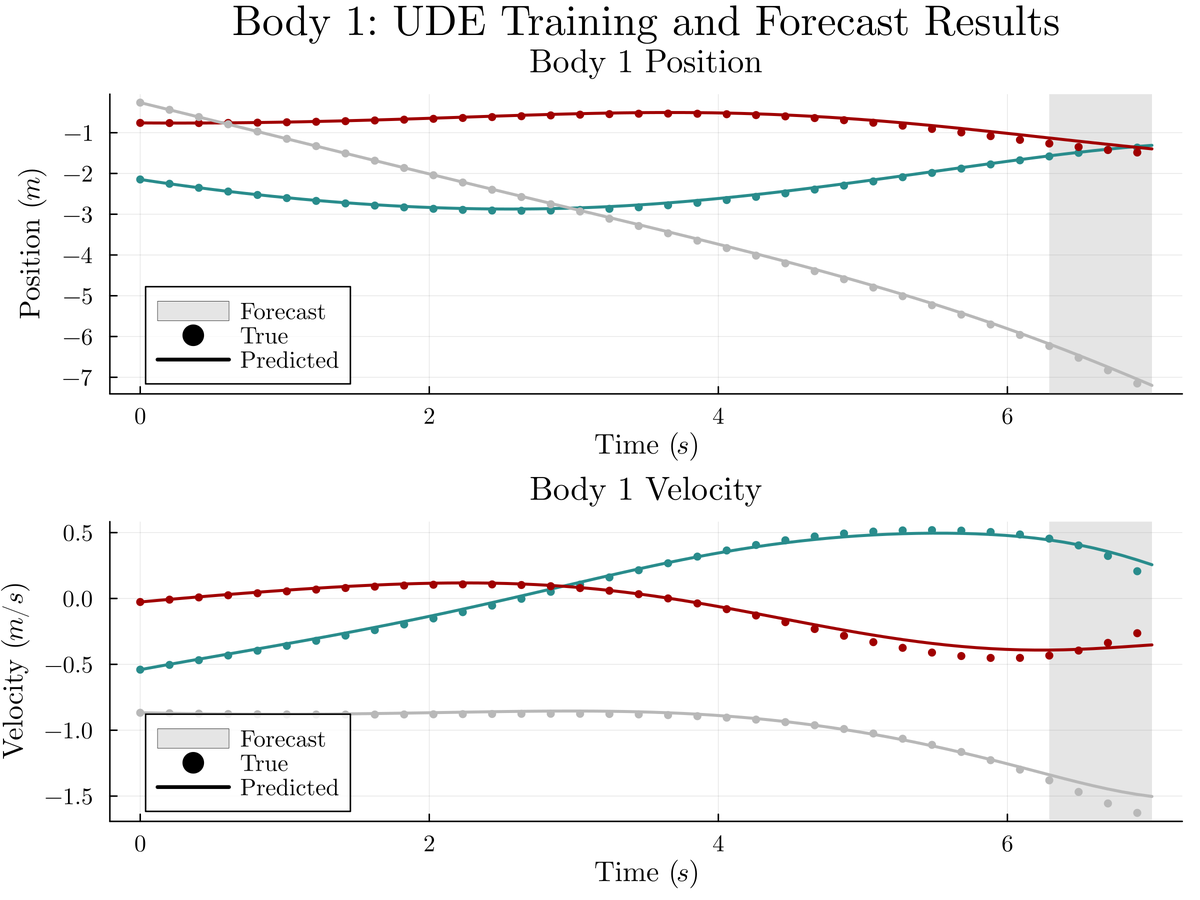}
        \caption{No noise}
    \end{subfigure}
    \hfill
    \begin{subfigure}[b]{0.32\columnwidth}
        \includegraphics[width=\linewidth]{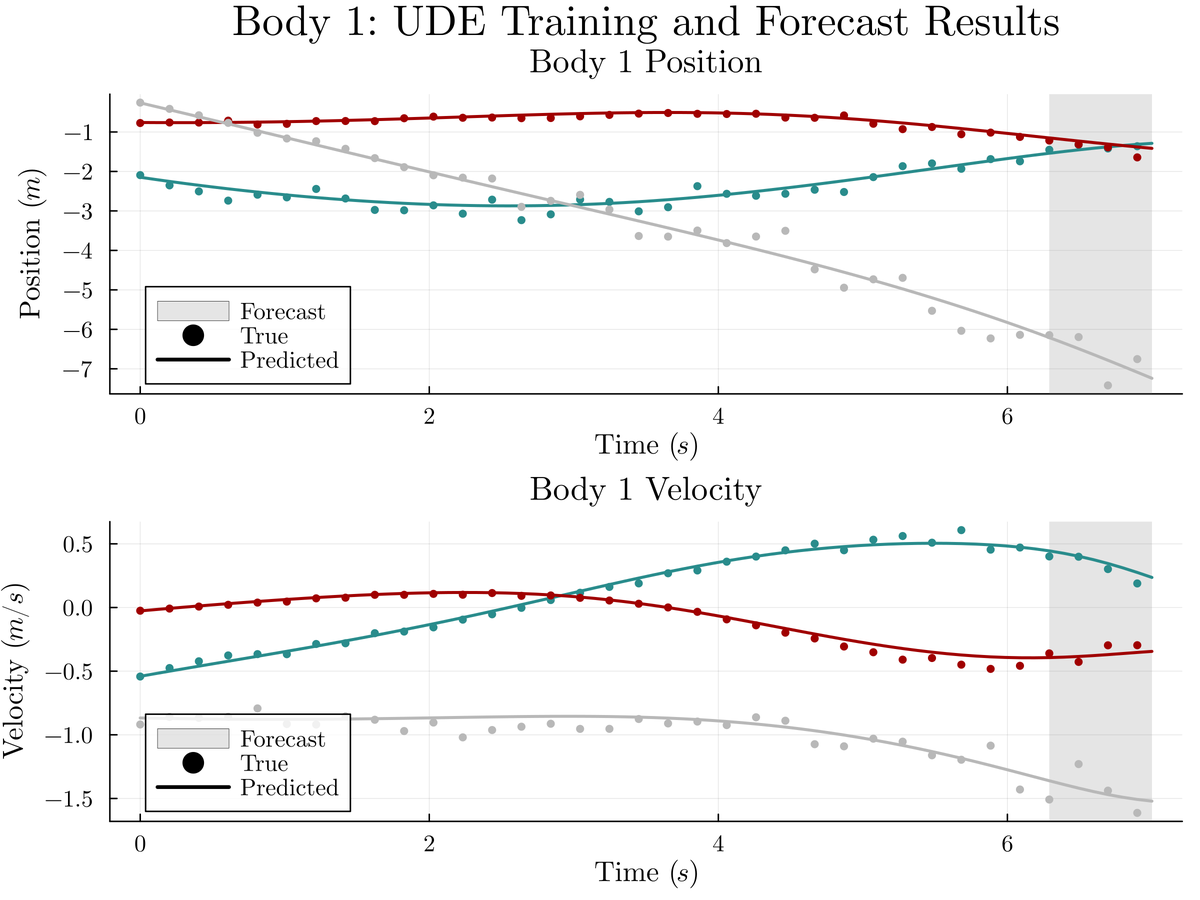}
        \caption{Moderate noise}
    \end{subfigure}
    \hfill
    \begin{subfigure}[b]{0.32\columnwidth}
        \includegraphics[width=\linewidth]{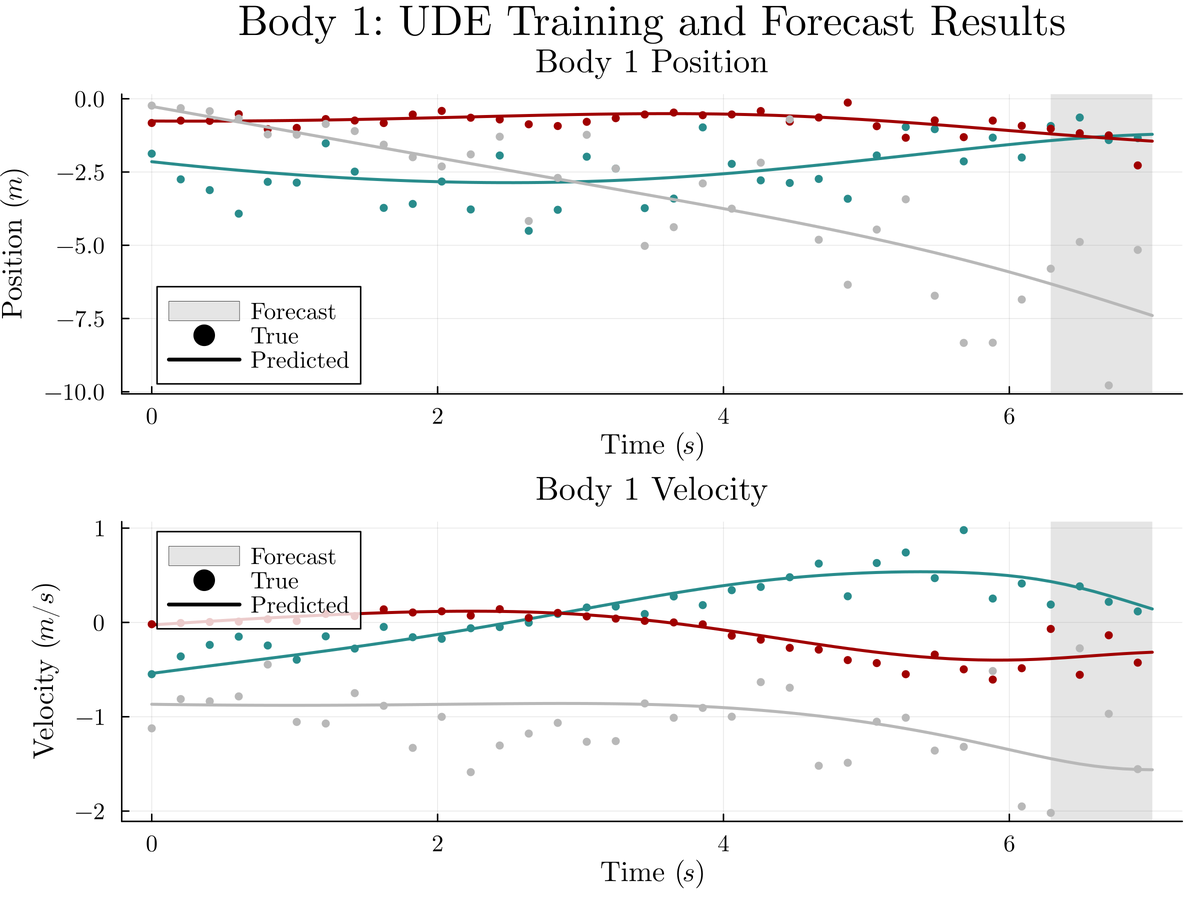}
        \caption{High noise}
    \end{subfigure}
    \caption{UDE results for Case 2 (90\% training) across different noise levels for body 1.}
    \label{fig:case2_ude_app}
\end{figure}

From Figure \ref{fig:case2_ude_app}, the UDE performs exceptionally well, as expected. With 90\% of the data, the model's predictions are virtually indistinguishable from the ground truth across all noise levels. It produces a clean, accurate forecast that perfectly captures the system's dynamics, showcasing its superior robustness and reliability when given ample data.

\subsubsection{Case 4: Training on 40\% of dataset and forecasting}
This case evaluates the models' performance when trained on 40\% of the data and tasked with forecasting the remaining 60\%. This scenario is well below the breakdown point for the Neural ODE but remains a viable test for the UDE.

\begin{figure}[h!]
    \centering
    \begin{subfigure}[b]{0.32\columnwidth}
        \includegraphics[width=\linewidth]{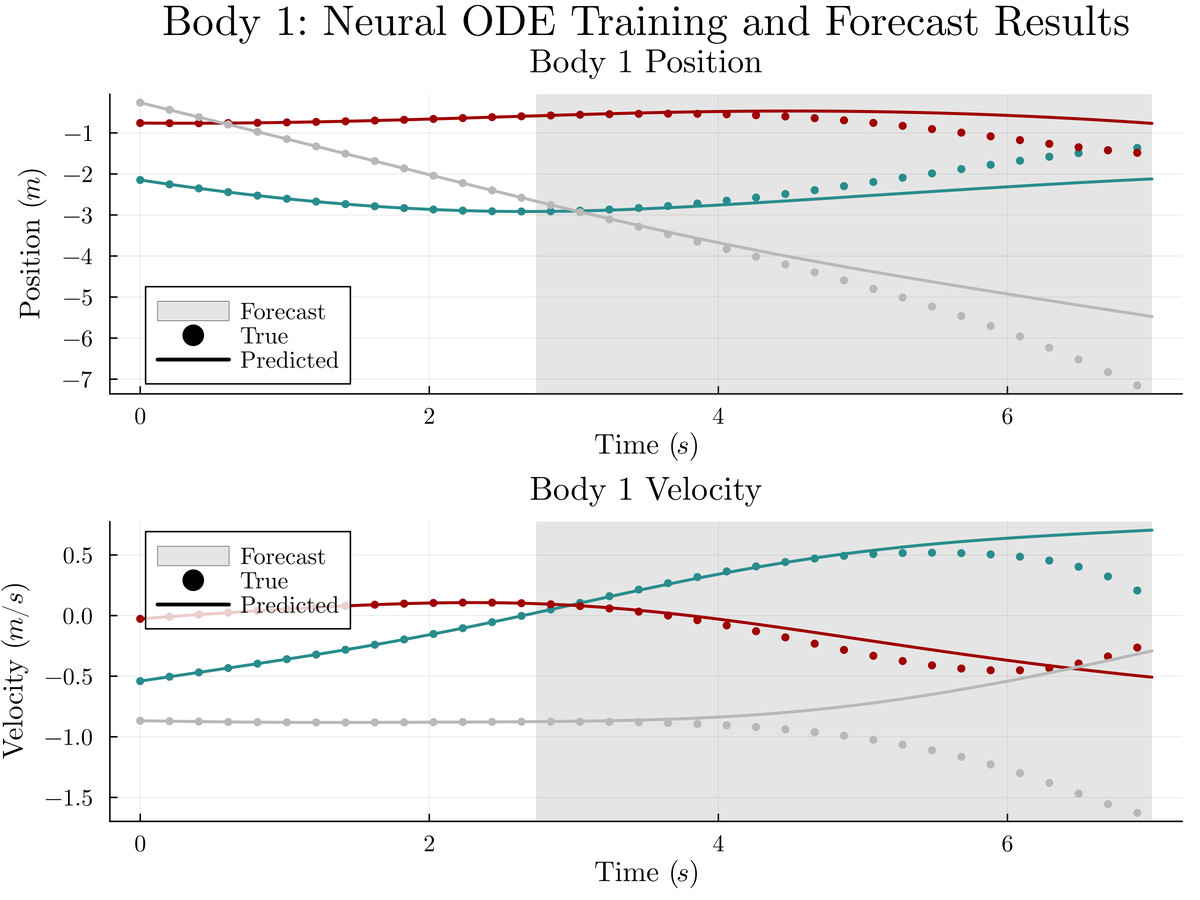}
        \caption{No noise}
    \end{subfigure}
    \hfill
    \begin{subfigure}[b]{0.32\columnwidth}
        \includegraphics[width=\linewidth]{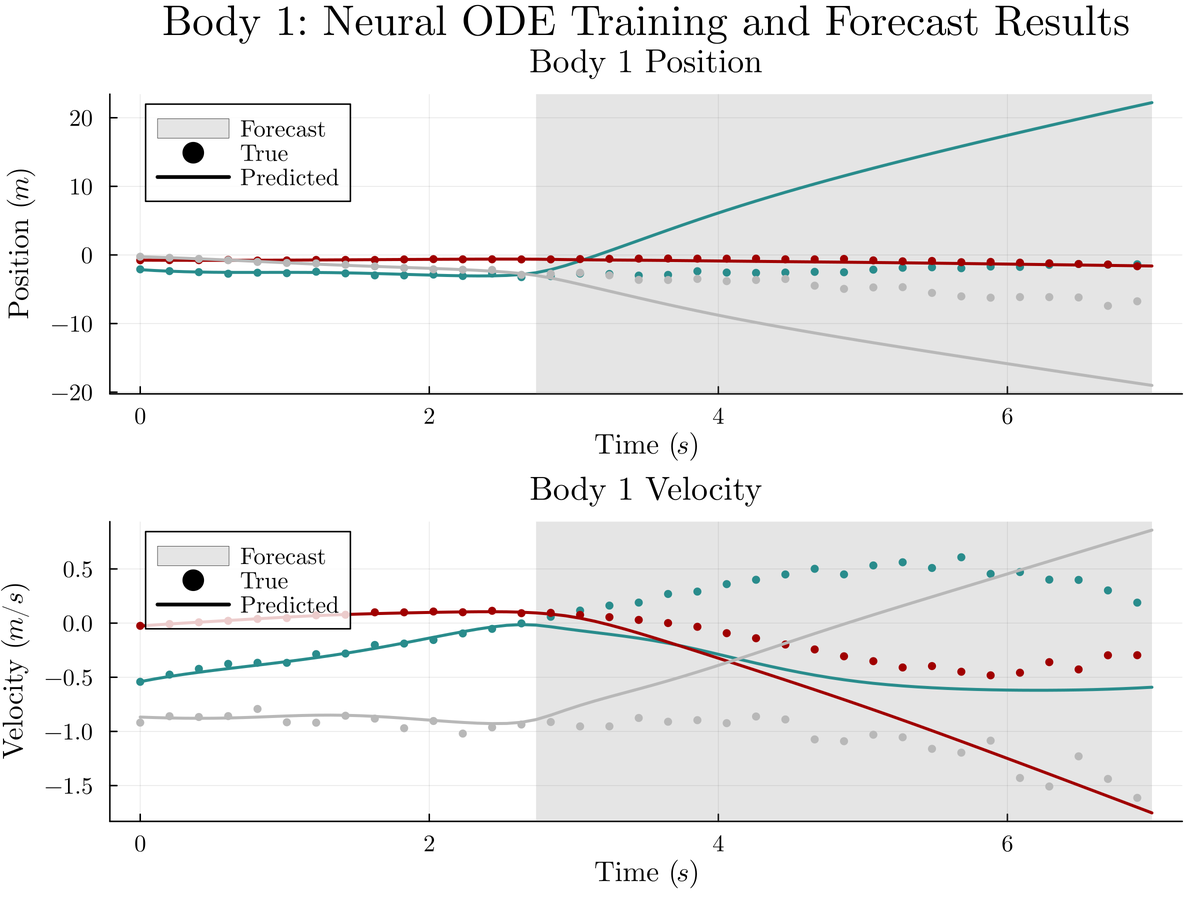}
        \caption{Moderate noise}
    \end{subfigure}
    \hfill
    \begin{subfigure}[b]{0.32\columnwidth}
        \includegraphics[width=\linewidth]{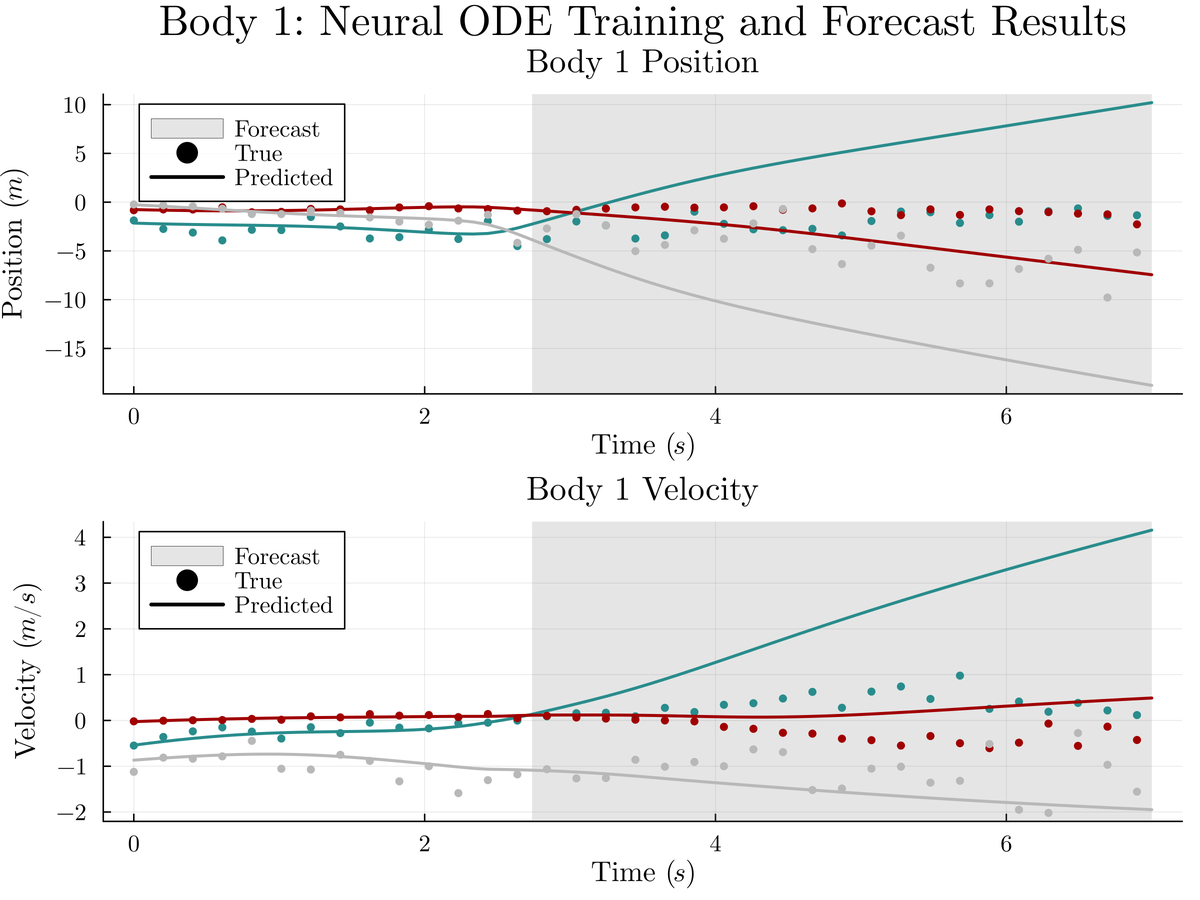}
        \caption{High noise}
    \end{subfigure}
    \caption{Neural ODE results for Case 4 (40\% training) across different noise levels for body 1.}
    \label{fig:case4_node_app}
\end{figure}

As illustrated in Figure \ref{fig:case4_node_app}, generalizability with scarce data remains a challenge for Neural ODE. Without noise, the model's prediction gradually follows that of the ground truth but fails at the long horizon, with its trajectory deviating largely. Such instability is continuously escalated by noise. For moderate noise, it deviates earlier and more distinctively, while for high-noise, it rapidly degenerates to a physically unreasonable trajectory. This verifies that 40\% data coverage proves incapable for the purely data-driven model to acquire the underlying dynamics, least of which under noisy conditions.

\begin{figure}[h!]
    \centering
    \begin{subfigure}[b]{0.32\columnwidth}
        \includegraphics[width=\linewidth]{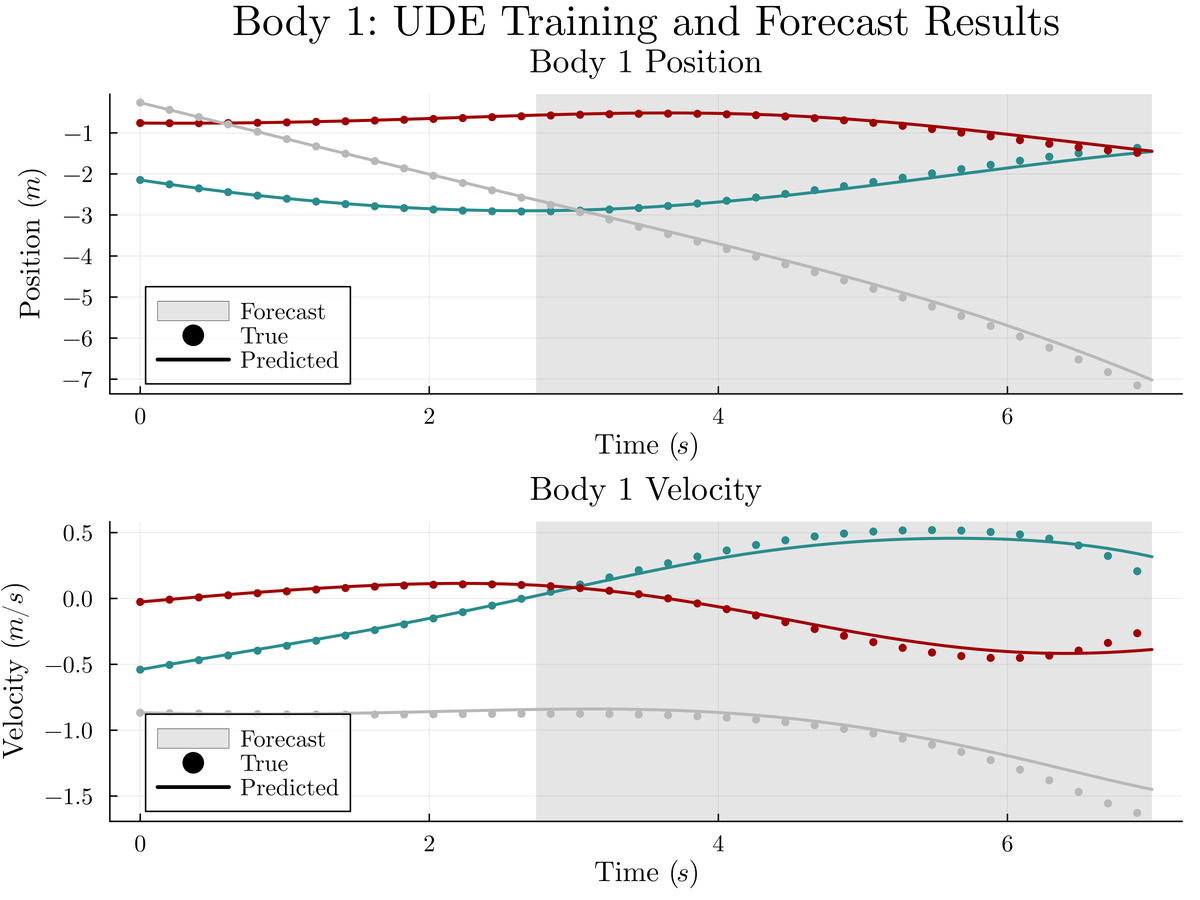}
        \caption{No noise}
    \end{subfigure}
    \hfill
    \begin{subfigure}[b]{0.32\columnwidth}
        \includegraphics[width=\linewidth]{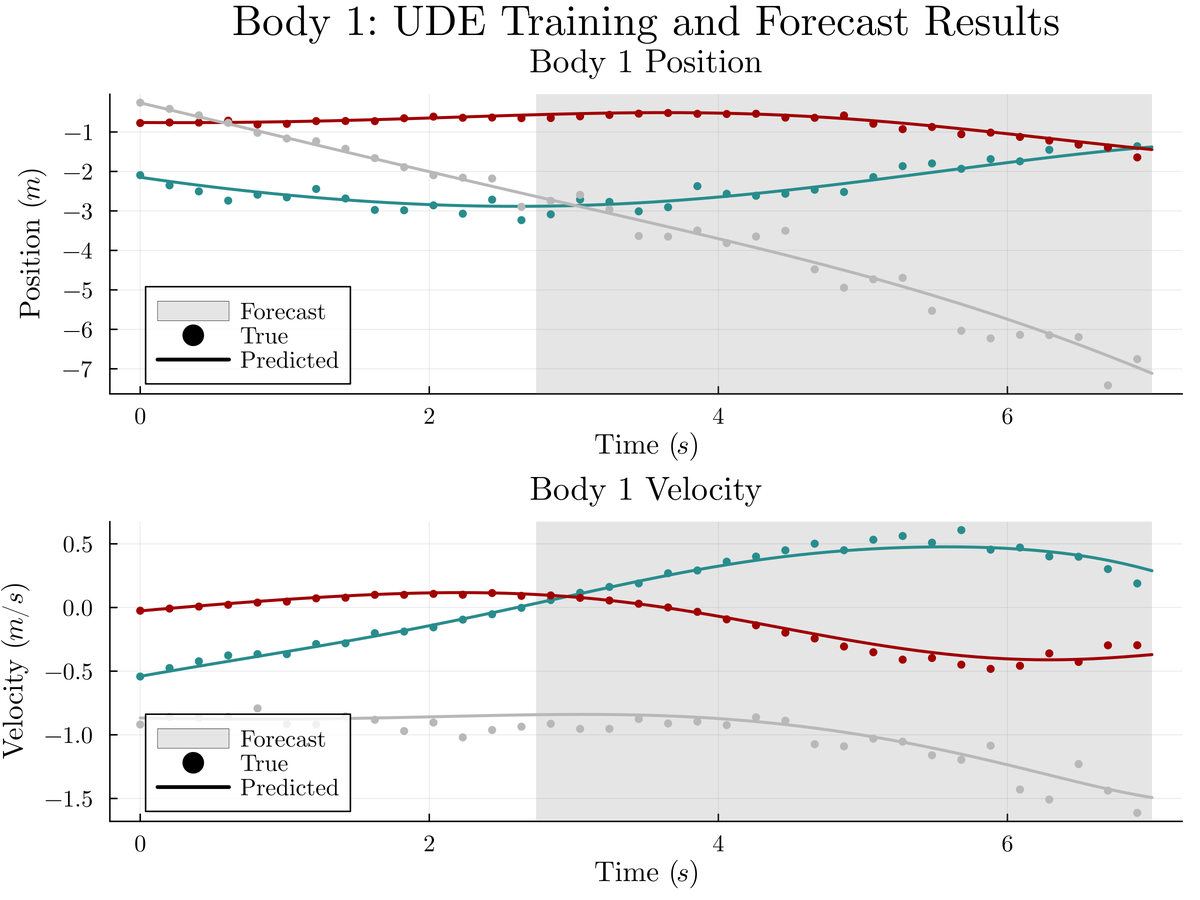}
        \caption{Moderate noise}
    \end{subfigure}
    \hfill
    \begin{subfigure}[b]{0.32\columnwidth}
        \includegraphics[width=\linewidth]{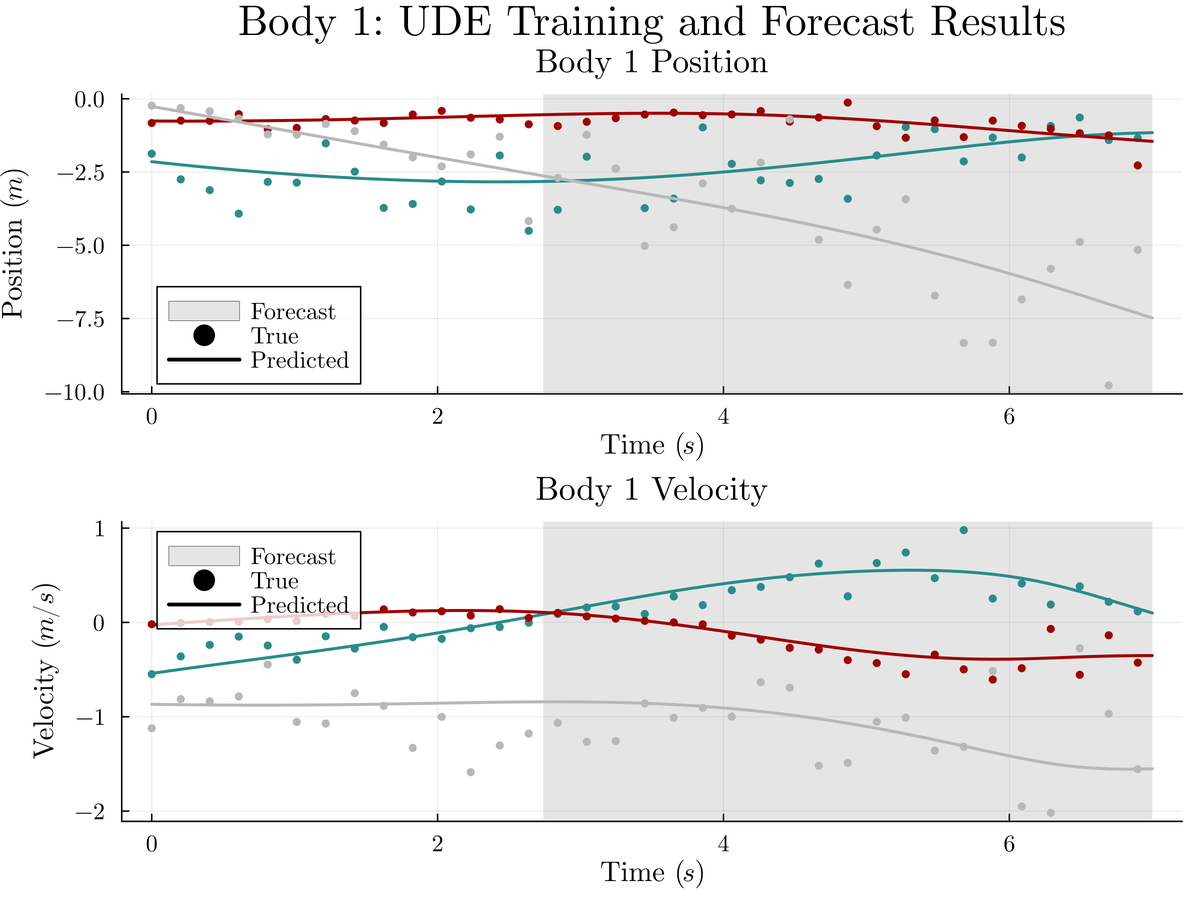}
        \caption{High noise}
    \end{subfigure}
    \caption{UDE results for Case 4 (40\% training) across different noise levels for body 1.}
    \label{fig:case4_ude_app}
\end{figure}

In contrast, Figure \ref{fig:case4_ude_app} demonstrates the UDE's continued robustness. The model provides a stable and accurate long-range forecast in the no-noise and moderate-noise conditions. Under high noise, some minor amplitude and phase errors accumulate over the 60\% forecast window, but the overall trajectory remains physically coherent and closely follows the underlying dynamics. This starkly contrasts with the Neural ODE's failure, highlighting the critical advantage of incorporating physical priors.

\end{document}